\documentclass[11pt]{article}

\usepackage[preprint]{acl}

\usepackage{times}
\usepackage{latexsym}
\usepackage[T1]{fontenc}
\usepackage[utf8]{inputenc}
\usepackage{microtype}
\IfFileExists{inconsolata.sty}{\usepackage{inconsolata}}{}

\usepackage{graphicx}
\usepackage{amsmath}
\usepackage{booktabs}
\usepackage{multirow}
\usepackage{subcaption}
\usepackage{placeins}
\usepackage{needspace}
\usepackage{tcolorbox}

\usepackage{enumitem}

\usepackage{amsmath,amsfonts,bm}

\def\Figref#1{Figure~\ref{#1}}

\def\eqref#1{equation~\ref{#1}}
\def\1{\bm{1}}

\DeclareMathAlphabet{\mathsfit}{\encodingdefault}{\sfdefault}{m}{sl}
\SetMathAlphabet{\mathsfit}{bold}{\encodingdefault}{\sfdefault}{bx}{n}

\usepackage[textsize=tiny]{todonotes}

\title{Vision-Language Models Integrate Conflicting Evidence Like Biased Bayesian Observers}

\author{
 Josh Barua\hspace{0.5cm}
 Kyle Mahowald\hspace{0.5cm}
 William Rudman
\\
The University of Texas at Austin
\\
\texttt{\{joshbarua,mahowald,william.rudman\}@utexas.edu}
}

\begin{document}
\maketitle

\begin{abstract}
% Vision-language models (VLMs) must make decisions under uncertainty when text and images conflict.
% Prior work has characterized the resulting behavior in terms of  modality biases—e.g., a ``blind faith in text''—but these accounts do not explain why preferences vary across domains, models, and input conditions.
% We develop a cue combination model from a Bayesian perspective, positing that VLMs integrate evidence independently across modalities and weight each source by its reliability. 
% Our model is predictive of VLM behavior and explains prior empirical observations.
Vision-language models must reconcile conflicting text and images.
Prior work has characterized the resulting behavior in terms of  modality biases—e.g., a ``blind faith in text''—but why preferences vary across models, domains, and input conditions remains unclear. 
We develop a predictive account from a Bayesian perspective, positing that VLMs combine conditionally independent evidence from each modality with prior beliefs, subject to modality-specific biases. 
The resulting two-parameter cue combination model accurately predicts VLM preferences under unseen combinations of input degradation and transfers to new forms of unreliability without refitting. 
Separating evidence strength from integration bias explains how VLMs can follow text even while weighting visual evidence more heavily, reconciling reports of text and vision dominance under a single combination rule.
\end{abstract}

\section{Introduction}\label{sec:introduction}

\begin{figure*}[t]
  \centering
  \includegraphics[width=0.9\textwidth]{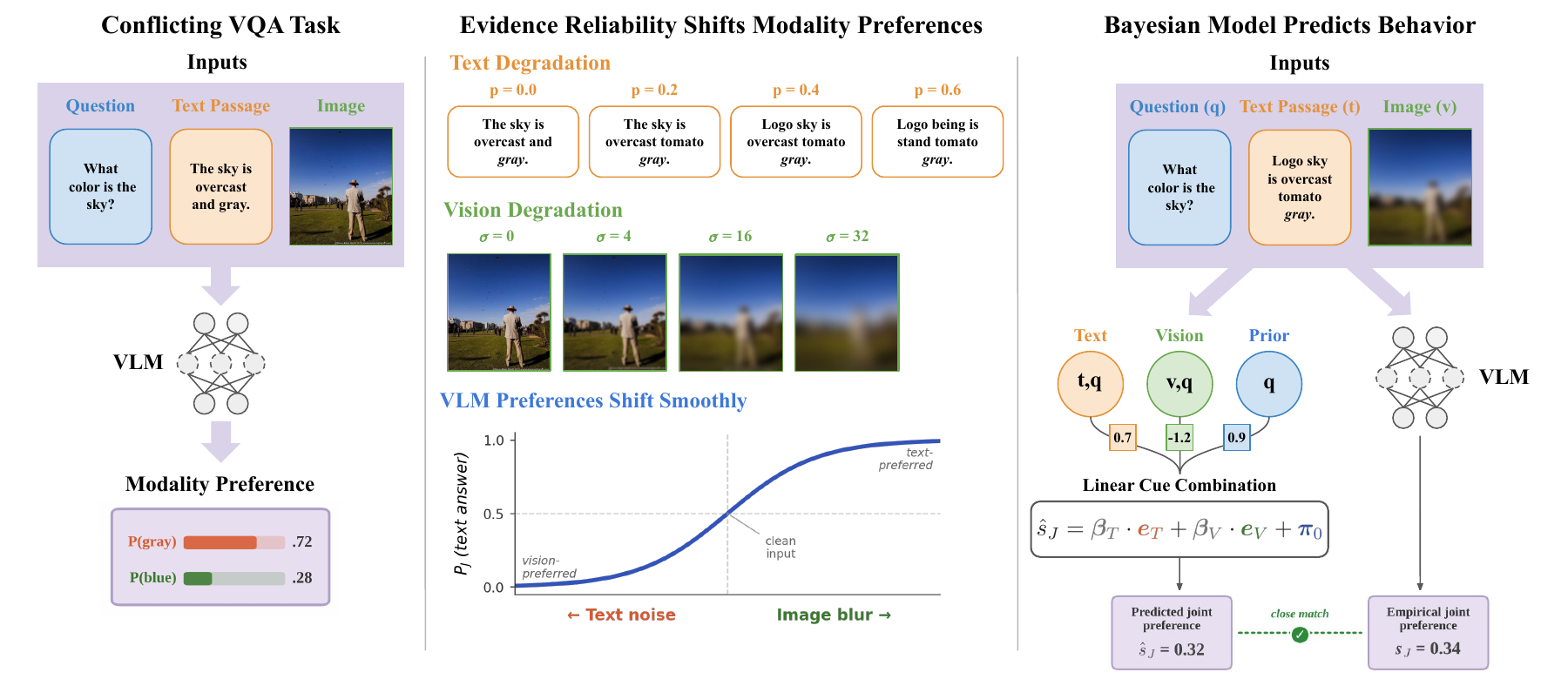}
  \caption{\textbf{Overview.} \textbf{(Left)} In the conflicting VQA task, a question is paired with a text passage and an image that support different answers ($a_T \neq a_V$); the VLM arbitrates between them and outputs a preference. \textbf{(Middle)} We independently degrade textual evidence (text noise level $p$) and visual evidence (Gaussian blur $\sigma$); as the reliability of each modality shifts, the VLM's preference transitions between the text- and vision-supported answers. \textbf{(Right)} A two-parameter Bayesian model over the unimodal Bayes factors ($e_T$, $e_V$) and the prior belief ($\pi_0$) accurately predicts the joint preference $s_J$ of the VLM.}
  \label{fig:overview}
\end{figure*}

Vision-language models (VLMs) are increasingly being deployed in high-stakes environments, such as web browsing \cite{he2024web} and medical diagnostics \cite{moor2023foundation}, where multimodal inputs may provide conflicting evidence. 
Such conflicts can arise from noisy sensors \cite{liao2025benchmarking}, adversarial attacks \cite{schlarmann2023adversarial, cui2024robustness}, human errors \cite{brady2017error}, or mismatches between evidence and parametric knowledge \cite{liu2025insight, golovanevsky2025pixels}.
To better understand how VLMs handle such conflicts, recent studies have focused on \textit{modality preference}, the extent to which a VLM follows one modality over another when the two disagree.
%Prior work often describes answering from the wrong source as ``hallucination'' or ``modality bias.''
%But in the presence of conflicting evidence, the model is also facing a more basic computational problem: it must arbitrate between two imperfect sources of information.
Early work frequently described VLM behavior in terms of particular biases.
\citet{deng2025blind} reported a ``blind faith in text'' where VLMs excessively trust text over vision when inconsistencies arise; others reported vision-leaning behavior \cite{yan2026beyond} or biases that vary with model architecture \cite{hua2025vision} and query complexity \cite{pezeshkpour2025mixed}.
Mechanistic studies reinforce this picture, finding that the modality dominating an answer can often be traced to specific internal components such as the attention mechanism \cite{hua2025vision, zheng2026malora, rudman2026mechanisms} or steering directions \cite{golovanevsky2025pixels, zhang2026evaluating, nooralahzadeh2026arbitration}. 
More recently, \citet{zhang2025modalities} showed that preference depends on both an inherent bias \textit{and} each modality's relative uncertainty, with predictions oscillating across layers when the two are balanced.

\smallskip
\noindent This broader trajectory closely mirrors one in the study of human perception.
Conflicts between senses were initially explained as one sense \textit{dominating} another \cite{rock1964vision}, as in the ventriloquism effect, where a speech sound can be perceived as coming from somewhere other than its true physical source if it matches visual cues \cite{howard1966human}.
%Experiments on these kinds of effects that varied cue reliability instead showed that which sense dominates depends on which is more \textit{reliable} \citep{ernst2002humans,alais2004ventriloquist}.
%Experiments that varied cue reliability instead showed that each sense's influence scales with its reliability \citep{ernst2002humans,alais2004ventriloquist}.
Later experiments that degraded visual cues instead showed that which sense dominates can change with cue \textit{reliability} \citep{heller1983haptic}.
%These findings can be explained by Bayesian accounts in which perception combines uncertain evidence from each sense with prior expectations \citep{battaglia2003bayesian, knill2004bayesian}.
These findings can be explained by Bayesian ideal-observer models in which each sense contributes evidence weighted by its reliability, so that influence shifts gradually rather than switching \citep{ernst2002humans, alais2004ventriloquist, knill2004bayesian}.
%The account extends to categorical judgments, where each source contributes a separate term to the log-odds of a response \citep{massaro1998perceiving,movellan2001morton}.
The framework also remained useful where people fell short of the ideal observer.
A systematic bias toward vision \citep{battaglia2003bayesian} and individual differences in how listeners weight sound and sight \citep{schwartz2010dependent} were captured by additional parameters within it.
Motivated by this history, we ask whether the same framework can usefully explain how VLMs arbitrate between text and images.

\smallskip
\noindent \textbf{This work.}
Following studies of human perception, we design a controlled setup to independently degrade the reliability of textual and visual evidence under conflict (\S\ref{sec:reliability}).
Consistent with \citet{zhang2025modalities}, VLM preferences shift smoothly toward whichever source is more reliable, and this pattern varies systematically across models and domains.
However, such shifts describe behavior without revealing the \textit{computational strategy} that produces it \cite{marr1982vision}.
A VLM that switches to the more confident modality and one that blends the two modalities could both produce the same behavior. 
Likewise, a VLM may follow the text because the passage supplies stronger evidence or because it weights textual evidence more heavily.
%Telling these apart requires a model that commits to an integration rule and can therefore predict preferences it was not fit to.
%We derive such a model from how a Bayesian observer would integrate evidence across modalities (\S\ref{sec:bayesian}).
To distinguish these possibilities, we derive how a Bayesian observer would integrate evidence across modalities (\S\ref{sec:bayesian}).
Under two assumptions, this yields a \textit{linear cue combination} model in which the VLM (i) treats each modality’s evidence as independent given the answer, (ii) weights it by a modality-specific bias, and (iii) combines it additively to its prior beliefs. 
Our model accurately predicts VLM preferences under unseen combinations of degradation and transfers to new forms of unreliability without refitting.
This predictive success lets us interpret the model's components as a computational-level theory of how VLMs arbitrate between modalities.
This recasts the ``blind faith in text'' \citep{deng2025blind} as largely a consequence of stronger textual evidence, which modality bias can amplify or offset, and reconciles reports of text and vision dominance under a single combination rule.
%More broadly, our work contributes to a growing body of literature applying Bayesian theories and methods to understand complex behavior in LLMs \cite{xie2022explanation, bigelow2026belief, vasileiou2026do}.

% overall introduction steps 
% 1. motivate the study of modality conflict with implications for decision-making in high stakes settings
% 2. cover prior works benchmarking modality preferences and attempting to control them 
% 3. conclude with zheng et al who first tie modality preference to relative reasoning uncertainty and inherent preference 
% 4. discuss how study of perceptual cues is not new and has long been studied in humans
% 5. this work 
% 6. conclude with how this extends bayesian accounts of explaining phenomena such as in context learning to multimodal conflict resolution

\section{Do Modality Preferences Track Evidence Reliability?}\label{sec:reliability}
Most existing modality-conflict benchmarks pair each question with clean text and image inputs that differ only in the answer they support \cite{hua2025vision, deng2025blind, pezeshkpour2025mixed, zhang2026evaluating}. 
These evaluations measure how often a VLM follows each modality, but because the evidence is held fixed, they are limited to coarse heuristics (e.g., a VLM chose the text answer 80\% of the time) and cannot show how preference depends on the certainty of the evidence itself. 
We therefore borrow the cue-conflict paradigm from human perception and manipulate the reliability of each modality’s evidence independently.

\begin{figure*}[t]
  \centering
  \includegraphics[width=0.9\textwidth]{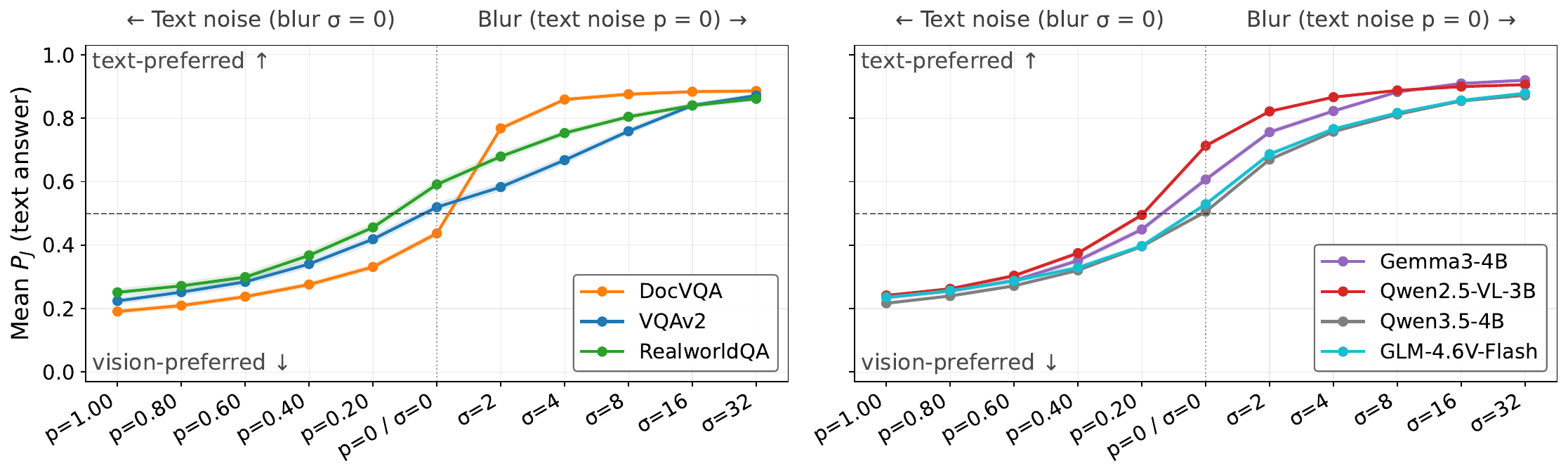}
  \caption{\textbf{Modality preferences shift sigmoidally with reliability and vary across datasets and VLMs.} Mean $P_J$ (probability of the text answer) as evidence from each modality is degraded. Within each panel, the left half is text noise $p$ with blur $\sigma{=}0$, the right half is blur $\sigma$ with text noise $p{=}0$, and the clean input $(p{=}0,\sigma{=}0)$ sits at the center. \textbf{Left:} per-dataset curves for Qwen3.5-4B. \textbf{Right:} per-model curves averaged over all datasets.}%Shaded bands show $\pm 1$ standard error of the mean across examples.}
  \label{fig:preference-curves}
\end{figure*}

\subsection{Task, Data, and Models}\label{sec:task-data-models}
Following prior work, our chosen setting is \textit{conflicting VQA}. 
Formally, each example is a tuple $(q, t, v, a_T, a_V)$ where $q$ is a question, $t$ is a text passage, $v$ is an image, $a_T$ is the answer supported by the text, and $a_V$ is the answer supported by the image, with $a_T \neq a_V$. 
In Figure~\ref{fig:overview} (left), the question is ``What color is the sky?''; the passage describes the sky as overcast and gray ($a_T = \textit{gray}$), while the image shows a clear blue sky ($a_V = \textit{blue}$).
Because neither answer is privileged as correct, this task isolates how a VLM arbitrates between sources from how accurately it answers.

\smallskip
\noindent We construct a dataset of 2{,}765 examples from three VQA benchmarks covering distinct visual domains: document images from DocVQA \cite{mathew2021doc}, natural photographs from VQAv2 \cite{antol2015vqa}, and real-world scenes from RealworldQA \cite{xai2024grok}.
%DocVQA \cite{mathew2021doc} consists of document images (forms, receipts, letters) where answering questions typically requires reading text rendered in the image. 
%VQAv2 \cite{antol2015vqa} contains natural photographs probing object recognition, counting, color, and spatial relations. 
%RealworldQA \cite{xai2024grok} consists of real-world scenes captured from vehicles and other naturalistic settings, testing spatial and contextual reasoning. 
% Figure~\ref{fig:dataset-examples} shows one example from each dataset.
For DocVQA and VQAv2, the conflicting passages are sourced from an existing benchmark \cite{deng2025blind}; for RealworldQA, we generate them with an LLM.
Further details are provided in Appendix~\ref{app:dataset-construction}.
% We further verify that each modality favors its intended answer in Table~\ref{tab:probe-sign-agreement}.

\smallskip
\noindent We evaluate four open-weight VLMs: Qwen2.5-VL-3B \cite{qwen2.5-VL}, Qwen3.5-4B \cite{qwen3.5}, Gemma3-4B \cite{Kamath2025Gemma3}, and GLM-4.6V-Flash \cite{glm45v}.
They differ in vision encoder, language-model attention, and vision--language fusion, which lets us determine whether our conclusions are consistent across architectures.
Appendix~\ref{app:models} lists the model checkpoints used in the experiments.

\subsection{Manipulating Evidence Reliability}\label{sec:reliability-manipulations} 
We introduce two corruptions that independently degrade the reliability of each modality (\Figref{fig:overview}, middle).
For images, we apply isotropic \textbf{Gaussian blur} with $\sigma \in \{0,2,4,8,16,32\}$ pixels, a standard controlled corruption in robustness studies \cite{borkar2019deepcorrect,hendrycks2019benchmarking}.
For text, we introduce \textbf{text noise} by randomly replacing a fraction $p \in \{0.0, 0.2, 0.4, 0.6, 0.8, 1.0\}$ of the words in the passage.
Replacement words are drawn from a vocabulary of that dataset's passages without regard to part of speech, and substitutions are nested: a word replaced at level $p$ stays replaced, by the same word, at every higher level.
This corruption is motivated by noisy-channel models of language comprehension, which treat observed language as a potentially corrupted realization of an intended message \cite{levy2008noisy,gibson2013rational,ryskin2018comprehenders}.
Under this view, out-of-place words and broken syntax are themselves evidence that a passage is unreliable.
Crossing the two corruptions yields a $6 \times 6$ grid of $(\sigma, p)$ conditions for every example.

\subsection{Measuring Modality Preference}\label{sec:modality-preference}
We define a VLM's \textit{modality preference} on a conflicting example as the log-odds between the text- and vision-supported answer, given the question and both sources of evidence:
\begin{equation}\label{eq:joint}
s_J = \log \frac{P(a_T \mid t, v, q)}{P(a_V \mid t, v, q)}.
\end{equation}
Throughout this work, a \textit{probe} is a behavioral measurement in the spirit of cloze-style probes of language model knowledge \cite{petroni2019language}. This differs from the more common usage, in which a probe is a classifier trained on a model's internal representations \cite{ettinger2016probing, alain2017understanding}.
We call $s_J$ the \textit{joint probe} because both modalities are in context, distinguishing it from the unimodal probes introduced in Section~\ref{sec:bayesian}.
Its sign indicates which answer the model prefers ($s_J > 0$ favors $a_T$; $s_J < 0$ favors $a_V$), and its magnitude indicates how strongly.
We score each candidate answer by its log-likelihood averaged over its tokens, so longer answers are not penalized.
For visualization, we convert $s_J$ to $P_J = 1/(1 + e^{-s_J})$, the probability of choosing $a_T$ over $a_V$.
Appendix~\ref{app:behavioral-measurements} gives the exact scoring definition and prompt templates and Appendix~\ref{app:open-ended-validation} verifies that the preference implied by the forced-choice log-odds largely agrees with the preference VLMs exhibit after reasoning.

\subsection{Results}\label{sec:reliability-results}
Figure~\ref{fig:preference-curves} shows the mean text-answer preference probability, $P_J$, as each modality is degraded.
Corrupting the passage shifts preference toward $a_V$, while blurring the image shifts it toward $a_T$.
This matches the trends observed by \citet{zhang2025modalities} when varying the difficulty of each modality under conflict.
Preferences thus respond systematically to changes in the available information, rather than remaining fixed on one modality.

\smallskip
\noindent These responses vary across both VLMs and datasets.
Qwen2.5-VL-3B and Gemma3-4B lean toward the text answer more than Qwen3.5-4B and GLM-4.6V-Flash (Fig.~\ref{fig:preference-curves}, right), consistent with earlier reports of text-following \citep{deng2025blind}.
Yet the direction of preference also changes within a single VLM.
On clean inputs, Qwen3.5-4B leans toward the image on DocVQA but toward the text on RealworldQA (Fig.~\ref{fig:preference-curves}, left).
Additionally, its sensitivity to degradation varies.
On DocVQA, mild blur ($\sigma{=}2$) is enough to move its preference from the image answer to the text answer, whereas the shift on the other datasets is more gradual.
Both patterns likely reflect DocVQA's reliance on fine-grained document details, which make a legible image highly informative but are lost under even mild blur.
A VLM's modality preference therefore depends on both the task and the degradation of its inputs.
The curves alone cannot tell whether a shift reflects stronger evidence, heavier weighting of that evidence, or the prior.
Next, we derive a Bayesian model that separates these contributions.

\section{A Bayesian Cue Combination Model}\label{sec:bayesian}
To explain why VLM preferences shift with the reliability of either modality, we turn to models of cue combination from human perception. 
When cues from different senses conflict, people weight each by its reliability, often closely matching an ideal Bayesian observer \cite{ernst2002humans, alais2004ventriloquist}.
In a categorical choice like ours, this weighting is implicit since a less reliable cue simply contributes weaker evidence \cite{massaro1998perceiving, movellan2001morton}. 
We formalize this account for VLMs, starting from a calibrated observer (\S\ref{sec:calibrated-observer}) and then relaxing it to allow systematic deviations (\S\ref{sec:integration-bias}), making each assumption explicit and testable.

\subsection{A Calibrated Bayesian Observer}\label{sec:calibrated-observer}
We treat the joint probe $s_J$ (Eq.~\ref{eq:joint}) as the posterior log-odds of an implicit observer.
Applying Bayes' rule decomposes $s_J$ into a Bayes factor for the evidence and the prior log-odds:
\begin{equation}\label{eq:joint-bayes-decomposition}
  s_J =
    \underbrace{\log \frac{P(t, v \mid q, a_T)}{P(t, v \mid q, a_V)}}_{\text{joint Bayes factor}}
    +
    \underbrace{\log \frac{P(a_T \mid q)}{P(a_V \mid q)}}_{\text{prior log-odds}}.
\end{equation}
We define the \textit{prior} probe $\pi_0 = \log \frac{P(a_T \mid q)}{P(a_V \mid q)}$, the model's baseline preference between the two answers without evidence.
We further decompose the joint Bayes factor into per-modality terms.\footnote{For the remainder of this section, we condition on the question~$q$ implicitly for brevity.}

\paragraph{Assumption~1 (Cross-modal conditional independence).}
Given the answer $A \in \{a_T, a_V\}$, the VLM treats the text and image as conditionally independent evidence: $P(t, v \mid A) = P(t \mid A) \cdot P(v \mid A)$.
This is the standard assumption in models of human cue integration, which treat cues from different senses as measurements of a common latent stimulus, each corrupted by independent sensory noise \citep{ernst2002humans, knill2004bayesian}.
% For a VLM, the assumption predicts that the joint probe contains no cross-modal interaction term.
Under this assumption, the joint Bayes factor splits additively:
\begin{equation}\label{eq:independent-bayes-factors}
  s_J =
    \underbrace{\log \frac{P(t \mid a_T)}{P(t \mid a_V)}}_{\text{text Bayes factor}}
    +
    \underbrace{\log \frac{P(v \mid a_T)}{P(v \mid a_V)}}_{\text{vision Bayes factor}}
    + \pi_0.
\end{equation}

\paragraph{From Bayes factors to observable probes.}
The per-modality Bayes factors in Eq.~\ref{eq:independent-bayes-factors} are not directly observable.
We can, however, measure the VLM's answer preference when only one modality is in context, using two \textit{unimodal} probes:
\begin{equation}\label{eq:unimodal-probes}
  s_T = \log \frac{P(a_T \mid t)}{P(a_V \mid t)}, \
  s_V = \log \frac{P(a_T \mid v)}{P(a_V \mid v)}.
\end{equation}
By Bayes' rule, each unimodal probe equals its Bayes factor plus the prior: $s_T = \log \frac{P(t \mid a_T)}{P(t \mid a_V)} + \pi_0$, and likewise for $s_V$.
For brevity, we define $e_T$ and $e_V$ as the text and vision Bayes factors, thus $s_T = e_T + \pi_0$ and $s_V = e_V + \pi_0$.
Finally, we rearrange and substitute $e_T$ and $e_V$ into Eq.~\ref{eq:independent-bayes-factors}, giving the \textit{calibrated naive-Bayes} prediction:
\begin{equation}\label{eq:calibrated-naive-bayes}
  s_J = e_T + e_V + \pi_0 = s_T + s_V - \pi_0.
\end{equation}
The observer is \textit{naive} because it treats text and image as independent evidence given the answer, and \textit{calibrated} because each modality shifts the log-odds as much in combination as alone.

\subsection{Allowing for Integration Bias}\label{sec:integration-bias}
Eq.~\ref{eq:calibrated-naive-bayes} gives each Bayes factor unit weight.
However, a VLM may systematically over- or underweight one modality relative to a calibrated observer.
Prior studies found that fitting human behavior required modeling such biases \cite{battaglia2003bayesian, schwartz2010dependent}.
% Observers were biased toward vision as a population on spatial localization tasks \cite{battaglia2003bayesian}, and audiovisual speech data required weights fit to each individual \cite{schwartz2010dependent}. 
We therefore relax the unit-weight assumption.

\paragraph{Assumption~2 (Modality-specific weighting).}
The VLM's multimodal decision combines the text and vision Bayes factors with modality-specific weights $\beta_T,\beta_V \geq 0$.
The weights are shared across examples, whereas evidence reliability still enters per example through $e_T$ and $e_V$.
Their ratio $\beta_T/\beta_V$ measures \textit{integration bias}, with values above one indicating greater weight on text evidence.

\paragraph{Linear Cue Combination Model.}
Applying these weights to the per-modality Bayes factors gives the final model shown in Figure~\ref{fig:overview} (right):
\begin{equation}\label{eq:cue_combination}
  \boxed{\;
  s_J
  =
  \beta_T e_T
  +
  \beta_V e_V
  +
  \pi_0.
  \;}
\end{equation}
This model decomposes a VLM's modality preference on each example into three sources: its prior between the candidate answers ($\pi_0$), the evidence each modality provides on that example ($e_T$, $e_V$), and the modality weights ($\beta_T$, $\beta_V$).

\subsection{Competing Predictors}\label{sec:predictors}
The linear cue combination model rests on implicit and explicit modeling choices: VLMs integrate both modalities rather than selecting one, evidence combines additively (Assumption 1), each modality carries its own weight (Assumption 2), and the prior enters with unit weight and no intercept.
We test each with a predictor that removes or relaxes it, mapping the same probes ($s_T$, $s_V$, $\pi_0$) to a prediction $\hat{s}_J$.
\begin{itemize}[leftmargin=*, itemsep=1pt]
    \item \textbf{Best Single Probe} follows the more confident modality: $\hat{s}_J = s_T$ if $|s_T| \geq |s_V|$, else $s_V$. This rule is reliability-sensitive and could in principle produce the shifts observed in \S\ref{sec:reliability}, but it ignores the weaker modality entirely. Our Bayesian model outperforming this probe is therefore evidence that VLMs blend information from both modalities rather than a winner-take-all strategy.
    \item \textbf{Bilinear Cue Combination} adds an interaction term to Eq.~\ref{eq:cue_combination}: $\hat{s}_J = \beta_T e_T + \beta_V e_V + \beta_I\, e_T e_V + \pi_0$. The effective weight on text evidence becomes $\beta_T + \beta_I e_V$, so the passage's contribution depends on how strongly the image favors an answer, and vice versa. Assumption~1 implies $\beta_I = 0$, so a substantial gain from the interaction term would count against it.
    \item \textbf{Calibrated Naive Bayes} is Eq.~\ref{eq:calibrated-naive-bayes}, the special case where $\beta_T = \beta_V = 1$. If it performed as well as the linear model, the modality weights (and therefore Assumption~2) would be deemed unnecessary.
    \item \textbf{Unconstrained Linear} fits $\hat{s}_J = \beta_T e_T + \beta_V e_V + \beta_\pi \pi_0 + \beta_0$, equivalent to linear regression on $s_T$, $s_V$, and $\pi_0$ with an intercept. Our Bayesian parameterization fixes $\beta_\pi = 1$ and $\beta_0 = 0$. Any gain over the Bayesian model measures the cost of these restrictions, while the fitted coefficients indicate departures from them.
\end{itemize}
Further details about the predictors can be found in Appendix~\ref{app:predictor-fitting}.

\section{Testing the Bayesian Model}\label{sec:bayesian-evaluation}
If linear cue combination is a useful computational-level account of modality preferences in VLMs, it should generalize beyond the conditions on which it was fit.
We therefore evaluate the Bayesian model on unseen combinations of degradation (\S\ref{sec:linear-prediction}), against competing integration rules (\S\ref{sec:predictors-results}), and on new reliability manipulations without refitting (\S\ref{sec:reliability-transfer}). 
If the model successfully generalizes, we may then interpret it and offer explanations for previously observed phenomena.

\begin{figure}[t]
  \centering
  \includegraphics[width=\columnwidth]{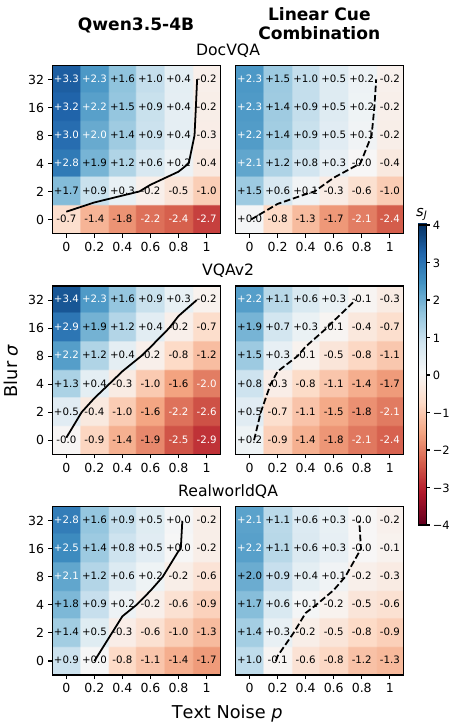}
  \caption{\textbf{Linear cue combination predicts the transition between text- and vision-dominated behavior.} Rows show DocVQA, VQAv2, and RealworldQA. Columns show empirical Qwen3.5-4B preferences (left) and model predictions (right), averaged over held-out examples under the OOD protocol. Solid and dashed lines mark the empirical ($s_J=0$) and predicted ($\hat{s}_J=0$) transition boundaries, respectively.}
  \label{fig:preference-surface-ood}
\end{figure}

\subsection{Fitting and Evaluation}\label{sec:evaluation-setup}
For each VLM and dataset, we fit the free parameters of the linear, bilinear, and unconstrained predictors by minimizing mean squared error (MSE) between $\hat s_J$ and $s_J$ with full-batch L-BFGS, using the validation set for early stopping.
The best single probe and calibrated naive Bayes have no fitted parameters.
Examples are split into train, validation, and test sets by example ID, so all $(\sigma,p)$ variants of an example belong to the same split.
We report test-set $R^2$ between predicted and observed joint preference scores, and KL divergence between the two-answer distributions induced by those scores.

\paragraph{Out-of-distribution protocol.}
Our main evaluation fits the parameters on the boundary of the degradation grid, $\{(\sigma,0)\}\cup\{(0,p)\}$ (11 cells), where at most one modality is degraded.
We then test on held-out examples in the 25 interior cells, where both modalities are degraded.
This protocol requires generalization to both new examples and unseen combinations of the same degradation levels, explicitly targeting the additive combination rule implied by Assumption~1.
Unless otherwise stated, evaluations in this section use this protocol.

\begin{table}[t]
  \centering
  \small
  \setlength{\tabcolsep}{4pt}
  \def\arraystretch{1.05}
  \begin{tabular}{p{5mm}lcc}
    \toprule
    & Predictor & KL $\downarrow$ & $R^2$ $\uparrow$ \\
    \midrule
    \multirow{5}{*}[-0.6em]{\rotatebox{90}{Qwen2.5}}
      & Best Single Probe       & 0.161 & 0.609 \\
      & Calibrated Naive Bayes       & 0.097 & 0.717 \\
      & Linear Cue Combination        & \textbf{0.057} & \textbf{0.852} \\
      \cmidrule(lr){2-4}
      & Bilinear Cue Combination      & 0.058 & 0.851 \\
      & Unconstrained Linear          & 0.058 & 0.852 \\
    \midrule
    \multirow{5}{*}[-0.6em]{\rotatebox{90}{Qwen3.5}}
      & Best Single Probe       & 0.240 & 0.452 \\
      & Calibrated Naive Bayes       & 0.137 & 0.477 \\
      & Linear Cue Combination        & \textbf{0.075} & \textbf{0.800} \\
      \cmidrule(lr){2-4}
      & Bilinear Cue Combination      & 0.075 & 0.801 \\
      & Unconstrained Linear          & 0.066 & 0.817 \\
    \midrule
    \multirow{5}{*}[-0.6em]{\rotatebox{90}{Gemma3}}
      & Best Single Probe & 0.787 & 0.472 \\
      & Calibrated Naive Bayes & 0.554 & 0.504 \\
      & Linear Cue Combination & \textbf{0.336} & \textbf{0.692} \\
      \cmidrule(lr){2-4}
      & Bilinear Cue Combination & 0.331 & 0.690 \\
      & Unconstrained Linear & 0.352 & 0.690 \\
    \midrule
    \multirow{5}{*}[-0.6em]{\rotatebox{90}{GLM-4.6V}}
      & Best Single Probe       & 0.417 & 0.592 \\
      & Calibrated Naive Bayes       & 0.268 & 0.625 \\
      & Linear Cue Combination        & \textbf{0.099} & \textbf{0.860} \\
      \cmidrule(lr){2-4}
      & Bilinear Cue Combination      & 0.099 & 0.860 \\
      & Unconstrained Linear          & 0.099 & 0.866 \\
    \bottomrule
  \end{tabular}
  \caption{\textbf{Goodness of fit of each predictor to VLM modality preferences under the OOD protocol.} $R^2$ is computed between the predicted $\hat{s}_J$ and observed $s_J$; KL is the divergence between the observed and predicted answer distributions. Metrics are averaged equally across DocVQA, VQAv2, and RealworldQA. Predictors above the inner rule have at most two free parameters ($\beta_T$, $\beta_V$). Predictors below it add expressivity: a cross-modal interaction term (bilinear) or a free prior coefficient and intercept (unconstrained linear). Bold marks the best predictor above the rule.}
  \label{tab:predictor-comparison}
\end{table}

\subsection{Predicting Joint Preferences from Unimodal Probes}\label{sec:linear-prediction}
As shown in Table~\ref{tab:predictor-comparison}, linear cue combination provides a good quantitative fit for VLM responses with $R^2$ of 0.80--0.86 for both Qwen models and GLM-4.6V, and 0.69 for Gemma3.
Figure~\ref{fig:preference-surface-ood} shows what this accuracy means at the level of the preference surface, using Qwen3.5-4B.
Averaged over held-out examples, the predictions reproduce the dataset-specific patterns from Section~\ref{sec:reliability}, including the abrupt shift toward text under mild blur on DocVQA, and the more gradual transitions on VQAv2 and RealworldQA.
On all three datasets, the predicted transition boundary ($\hat{s}_J=0$) closely tracks the observed one.
The same agreement holds for the other three VLMs we study (Appendix~\ref{app:ood-preference-grids}), and for larger models from the same families (Appendix~\ref{app:larger-models}).
Thus, unimodal preferences measured independently surprisingly capture most of the variation in VLMs preferences when both modalities are present and in conflict.
Gemma3's lower fit shows that the approximation is less complete for some models.

\begin{table}[t]
  \centering
  \small
  \setlength{\tabcolsep}{5pt}
  \def\arraystretch{1.1}
  \begin{tabular}{p{5mm}llc}
    \toprule
    & Fit Grid & Test Grid & $R^2$ \\
    \midrule
    \multirow{4}{*}[-0.6em]{\rotatebox{90}{Qwen2.5}}
      & relevance & relevance & \multirow{2}{*}{0.850} \\
      & $\times$ occlusion & $\times$ occlusion & \\
      \cmidrule(lr){2-4}
      & text noise & relevance & \multirow{2}{*}{0.852} \\
      & $\times$ blur & $\times$ occlusion & \\
    \midrule
    \multirow{4}{*}[-0.6em]{\rotatebox{90}{Qwen3.5}}
      & relevance & relevance & \multirow{2}{*}{0.806} \\
      & $\times$ occlusion & $\times$ occlusion & \\
      \cmidrule(lr){2-4}
      & text noise & relevance & \multirow{2}{*}{0.785} \\
      & $\times$ blur & $\times$ occlusion & \\
    \midrule
    \multirow{4}{*}[-0.6em]{\rotatebox{90}{Gemma3}}
      & relevance & relevance & \multirow{2}{*}{0.676} \\
      & $\times$ occlusion & $\times$ occlusion & \\
      \cmidrule(lr){2-4}
      & text noise & relevance & \multirow{2}{*}{0.683} \\
      & $\times$ blur & $\times$ occlusion & \\
    \midrule
    \multirow{4}{*}[-0.6em]{\rotatebox{90}{GLM-4.6V}}
      & relevance & relevance & \multirow{2}{*}{0.828} \\
      & $\times$ occlusion & $\times$ occlusion & \\
      \cmidrule(lr){2-4}
      & text noise & relevance & \multirow{2}{*}{0.832} \\
      & $\times$ blur & $\times$ occlusion & \\
    \bottomrule
  \end{tabular}
  \caption{\textbf{Linear cue combination and its weights generalize across reliability manipulations.} $R^2$ on joint log-odds, averaged equally across datasets. For each VLM, the first row refits weights on the new grid; the second tests generalization by transferring weights from text noise $\times$ blur without refitting. Both use the OOD protocol and identical test examples on relevance $\times$ occlusion.}
  \label{tab:degradation-transfer}
\end{table}

\begin{table*}[t]
  \centering
  \small
  \setlength{\tabcolsep}{4pt}
  \begin{tabular}{@{}llccc@{}}
    \toprule
    & & \multicolumn{2}{c}{$\beta_T/\beta_V$ [95\% CI]} & $|\bar e_T|/|\bar e_V|$ \\
    \cmidrule(lr){3-4}\cmidrule(lr){5-5}
    Model & Dataset & Example & Cell & Clean \\
    \midrule
    \multirow{3}{*}{Qwen2.5-VL} & DocVQA & $1.063_{[1.027, 1.102]}$ & $0.988_{[0.951, 1.025]}$ & 1.36 \\
     & VQAv2 & $1.449_{[1.375, 1.514]}$ & $1.273_{[1.209, 1.336]}$ & 1.40 \\
     & RealworldQA & $1.436_{[1.311, 1.519]}$ & $1.092_{[0.991, 1.187]}$ & 2.66 \\
    \midrule
    \multirow{3}{*}{Qwen3.5} & DocVQA & $0.661_{[0.632, 0.699]}$ & $0.651_{[0.619, 0.683]}$ & 1.44 \\
     & VQAv2 & $0.563_{[0.539, 0.626]}$ & $0.580_{[0.551, 0.609]}$ & 2.03 \\
     & RealworldQA & $0.675_{[0.635, 0.732]}$ & $0.630_{[0.588, 0.672]}$ & 2.88 \\
    \midrule
    \multirow{3}{*}{Gemma3} & DocVQA & $0.964_{[0.915, 1.017]}$ & $0.888_{[0.840, 0.940]}$ & 1.58 \\
     & VQAv2 & $1.044_{[0.973, 1.108]}$ & $1.028_{[0.963, 1.098]}$ & 1.41 \\
     & RealworldQA & $1.267_{[1.143, 1.386]}$ & $0.947_{[0.839, 1.065]}$ & 3.77 \\
    \midrule
    \multirow{3}{*}{GLM-4.6V} & DocVQA & $0.650_{[0.630, 0.692]}$ & $0.638_{[0.612, 0.664]}$ & 1.86 \\
     & VQAv2 & $0.620_{[0.595, 0.678]}$ & $0.609_{[0.574, 0.648]}$ & 2.83 \\
     & RealworldQA & $0.558_{[0.530, 0.626]}$ & $0.534_{[0.495, 0.575]}$ & 3.07 \\
    \bottomrule
  \end{tabular}
  \caption{\textbf{Evidence ratio and integration bias.} Integration bias above one indicates greater weighting of text evidence in the fitted rule. An evidence ratio above one indicates a larger magnitude of mean text evidence. Example-level fits predict individual scores; cell-level fits predict cell means, and serve as a robustness check. Bracketed subscripts give pointwise 95\% percentile intervals (Appendix~\ref{app:weight-bootstrap}). The evidence ratio uses the absolute values of signed mean evidence on clean held-out inputs ($\sigma=0$, $p=0$), with absolute values taken after averaging.}
  \label{tab:bias-decomposition}
\end{table*}

\subsection{Comparing Alternative Explanations}\label{sec:predictors-results}
A strong fit alone does not establish that linear cue combination has the right functional form.
Table~\ref{tab:predictor-comparison} compares it with the competing predictors (\S\ref{sec:predictors}), each configuration provides a different account of how a VLM could produce its joint preferences from the same probes.
The best single probe trails by 0.22--0.35 in $R^2$, which supports the argument that VLMs integrate evidence rather than deferring to the more confident modality.
Calibrated naive Bayes trails by 0.14--0.32, showing that fitting the modality weights substantially improves prediction over combining the evidence with no weights.
Additional flexibility yields much smaller gains.
Adding the bilinear interaction improves $R^2$ by at most 0.001, while unconstrained linear regression improves it by at most 0.017.
Although attention can mix the two modalities at any layer, and indeed does \cite{kaduri2025image, zhang2025cross}, we find that joint preferences are well approximated by an additive function of the unimodal evidence, as Assumption~1 predicts.
The unconstrained regression also tests the restriction Bayes' rule places on the prior.
For three of the four VLMs, its fitted prior coefficient is close to one (Table~\ref{tab:other-predictor-weights-ood}). %, so the constraint is recovered from the data rather than imposed.
The exception is Qwen3.5 ($\beta_\pi = 0.53$--$0.66$), whose question-only prior contributes less to joint preferences than Bayes' rule implies.
Thus, enforcing the Bayesian constraint costs little predictive accuracy.

\subsection{Do Modality Weights Transfer?}\label{sec:reliability-transfer}
Linear cue combination separates evidence, which the probes measure separately for each input, from modality weights, which are fixed for each VLM and dataset.
A challenging test of this separation is whether weights fit under one set of reliability manipulations still predict preferences under another.
We introduce two new manipulations that alter the evidence differently from the original corruptions (\S\ref{sec:reliability-manipulations}).
\textbf{Pixel occlusion} blacks out a growing fraction of the image, removing information locally while leaving the remaining pixels intact.
\textbf{Textual relevance} weakens a passage's support for the answer while preserving fluency and plausibility (details in Appendix~\ref{app:transfer-manipulations}).
After measuring the probes on the new relevance $\times$ occlusion grid, we compare two sets of weights, each evaluated with the OOD protocol on the same held-out examples.
\textit{Transferred} weights are fit on text noise $\times$ blur and applied without refitting to the new grid.
\textit{Refitted} weights are fit on the relevance $\times$ occlusion grid and serve as a reference.
If the weights are stable, the transferred weights should predict as well as the refitted ones, and the two sets of weights should be similar.

\smallskip
\noindent Strikingly, Table~\ref{tab:degradation-transfer} shows that both predictions hold.
Refitted weights reach $R^2$ of 0.68--0.85, and transferred weights come within 0.021 of the refit value for every VLM.
The refitted weights are also close to the transferred ones (App.~Table~\ref{tab:transfer-refit-weights}), so the new manipulations recover nearly the same modality weights.
This supports a key prediction of the model and suggests that the fitted modality weights are stable within each VLM--dataset pair.

\subsection{Explaining Modality Preferences}\label{sec:weights-biases}
Having shown that linear cue combination predicts joint preferences and that its weights transfer across reliability manipulations, we now turn to the interpretation it offers.
Under the rule $\hat s_J=\pi_0+\beta_Te_T+\beta_Ve_V$, a VLM's modality preference reflects the evidence each modality supplies, the weight assigned to that evidence, and the prior.
Table~\ref{tab:bias-decomposition} summarizes the first two factors as the clean-input evidence ratio, $|\bar e_T|/|\bar e_V|$, and the integration bias, $\beta_T/\beta_V$.

\smallskip
\noindent On clean inputs, mean textual evidence exceeds mean visual evidence by a factor of 1.4--3.8 for every VLM on every dataset.
This is expected given how the stimuli are constructed: the text states the answer while the image must be interpreted, though the ratio also depends on each model's ability to extract this information.
Stronger textual evidence can therefore lead a VLM to follow the text even with equal modality weights, unless the prior offsets this advantage.
Gemma3 illustrates this: it leans toward text more than Qwen3.5 and GLM-4.6V (Fig.~\ref{fig:preference-curves}, right), yet weights the two modalities nearly equally or modestly favors text ($\beta_T/\beta_V=0.96$--$1.27$).
Qwen3.5 and GLM-4.6V instead weight visual evidence more heavily on every dataset ($\beta_T/\beta_V=0.56$--$0.68$), counteracting the text advantage.
This vision-favoring weighting holds under both example- and cell-level fitting, whereas some smaller integration biases for Qwen2.5-VL and Gemma3 depend on the fitting level.
Strong text-following can thus coexist with only modest integration bias.
By the same logic, the ``blind faith in text'' reported by \citet{deng2025blind} and the visual preference reported in other models \citep{hua2025vision, yan2026beyond} do not necessarily reflect different integration strategies.
Under a single additive rule, which modality wins depends on how much evidence each supplies and how heavily it is weighted, and both vary across models and tasks.
Consistent with this, \citet{deng2025blind} also find that vision wins when it is the more certain modality.

\smallskip
\noindent The same decomposition explains how a single VLM's preference changes across datasets.
Qwen3.5 weights vision more heavily on all three datasets, but its evidence ratio rises from 1.44 on DocVQA to 2.03 on VQAv2 and 2.88 on RealworldQA.
The magnitude ratio of its weighted mean text and vision contributions correspondingly rises from 0.95 to 1.14 and 1.94, so the growing textual advantage increasingly outweighs its vision-favoring bias, qualitatively matching its increasing text preference (Fig.~\ref{fig:preference-curves}, left).
More generally, changes in evidence strength can reverse a VLM's preference even under fixed modality weights.
For example, \citet{pezeshkpour2025mixed} observe that VLM preferences shift toward images as textual queries grow more complex.
Our decomposition separates two explanations for this shift: weaker textual evidence under fixed weights, or a genuine change in integration bias.
Measuring $e_T$ and $e_V$ across query complexity would distinguish them.
\section{Conclusion}\label{sec:conclusion}
We presented a Bayesian cue combination model that predicts how VLMs arbitrate between conflicting text and images.
Despite its simple, two-parameter form, the model predicts per-example joint preferences across datasets, architectures, and held-out combinations of degradation, and its fitted weights transfer to new forms of unreliability without refitting.
Together, these results characterize modality preference as a predictable interaction between evidence strength and stable, modality-specific integration biases.
This distinction explains how apparent ``blind faith in text'' can arise even when visual evidence receives greater weight, reconciling reports of text and vision dominance under a single combination rule.

% \newpage

\section*{Limitations}
Our central claim is not that VLMs behave optimally or in a ``perfectly rational'' way, but that a Bayesian cue combination framework usefully explains their behavior. 
In particular, the modality weights $\beta_T$ and $\beta_V$ are fitted empirically rather than derived from a normative account, because models differ in how they weight text and vision evidence in ways our framework does not yet predict.
Further, we measure modality preferences using differences in length-normalized candidate-answer log-likelihoods.
Their correspondence to the posterior log-odds in our Bayesian derivation is approximate, and they do not directly measure open-ended generation frequencies.
%Because our Bayesian framework is formulated over probabilities, we must estimate the likelihood of each candidate answer.
%We measure modality preferences via forced-choice log-odds rather than open-ended generation, which would require many samples per condition and careful answer parsing. 
Although this is an approximation, Appendix~\ref{app:open-ended-validation} shows good agreement when candidate answers are given as multiple choice options and a VLM reasons over them.
Finally, we extensively manipulate the supplied evidence but do not systematically manipulate prior beliefs independently of that evidence.
Conflicts between supplied evidence and parametric knowledge offer a complementary setting in which to test the prior's contribution \cite{golovanevsky2025pixels, ortu2026seeing} .
Applying the framework to such counterfactual benchmarks would assess whether the elicited prior and measured evidence also account for preferences when external information conflicts with a model's existing knowledge.

% \section*{Acknowledgments}

% TODO: acknowledgments (omit in the review version).

% Bibliography. Use `\bibliography{anthology,custom}` once anthology.bib is
% downloaded into latex/ (see anthology.bib.txt for instructions).
\bibliography{custom}

\onecolumn
\appendix

% Appendix sections are kept in reader order; clear pages flush floats between major topics.
\clearpage
\section{Data and Reliability Manipulations}\label{app:data-manipulations}\label{app:experimental-details}

\FloatBarrier
\subsection{Dataset Construction}\label{app:dataset-construction}
Two procedures generate the conflicting tuples $(q, t, v, a_T, a_V)$, depending on the source benchmark. 
Both dataset-construction procedures use Gemini~3 Flash (\texttt{gemini-3-flash-preview}; \citealt{gemini3flash2025}), with task-specific generation settings given below.

\paragraph{Example counts.}
Table~\ref{tab:dataset-splits} reports the underlying example counts and split sizes; the partitioning procedure is specified in Appendix~\ref{app:evaluation-protocols}.

\begin{table}[!htbp]
  \centering
  \small
  \setlength{\tabcolsep}{5pt}
  \begin{tabular}{lrrrr}
    \toprule
    Dataset & Total & Train & Validation & Test \\
    \midrule
    DocVQA & 1{,}000 & 700 & 150 & 150 \\
    VQAv2 & 1{,}000 & 700 & 150 & 150 \\
    RealworldQA & 765 & 535 & 114 & 116 \\
    \midrule
    Total & 2{,}765 & 1{,}935 & 414 & 416 \\
    \bottomrule
  \end{tabular}
  \caption{Underlying example counts and split sizes for the per-dataset predictor comparisons. Counts exclude repeated degradation variants. Under the ID protocol, every example contributes $36$ cells; under the OOD protocol, training and validation examples contribute $11$ boundary cells each, and test examples contribute $25$ interior cells each.}
  \label{tab:dataset-splits}
\end{table}

\paragraph{DocVQA and VQAv2.} Source examples are drawn from \citet{deng2025blind}, which already pairs each $(q, v, a_V)$ from the original benchmarks with a programmatically corrupted text passage $t$ that is designed to suggest a different answer. 
We use Gemini~3 Flash to extract $a_T$ only: given $(q, a_V, t)$, the model is asked to report the answer that the text alone would suggest, constrained to a single word or short phrase in the same format as $a_V$.
For this extraction, we use temperature $0$, an output limit of $64$ tokens, and a thinking budget of $0$ tokens.

\paragraph{RealworldQA.} RealworldQA \citep{xai2024grok} examples are tuples $(q, v, a_V)$ in which many questions are multiple-choice or yes/no.
Thus, we must rewrite $q$ to be an open-ended question, restate $a_V$ as a word or short phrase (rather than an option letter), and generate a conflicting text passage $t$ that supports a different answer $a_T$.
Again, we use Gemini~3 Flash for all three of these tasks.
The generator receives the original image, question, and answer; the prompt explicitly permits substantial question reframing and asks it to state the image-supported answer to the rewritten question.
Figure~\ref{fig:realworldqa-generation-prompt} reproduces the full prompt.
For this RealworldQA conversion, we use temperature $0.7$, an output limit of $512$ tokens, and a thinking budget of $1{,}024$ tokens.
%The output must contain four nonempty JSON fields, and the two answers must differ after stripping whitespace and ignoring case; failed calls or invalid outputs are retried up to three attempts in total.
%The saved conversion cache contains valid outputs for all $765$ source examples.

\FloatBarrier
\Needspace{.55\textheight}
\subsection{Dataset Examples}\label{app:dataset-examples}

\begin{figure*}[!htbp]
  \centering
  \begin{subfigure}[t]{0.31\textwidth}
    \includegraphics[width=\textwidth]{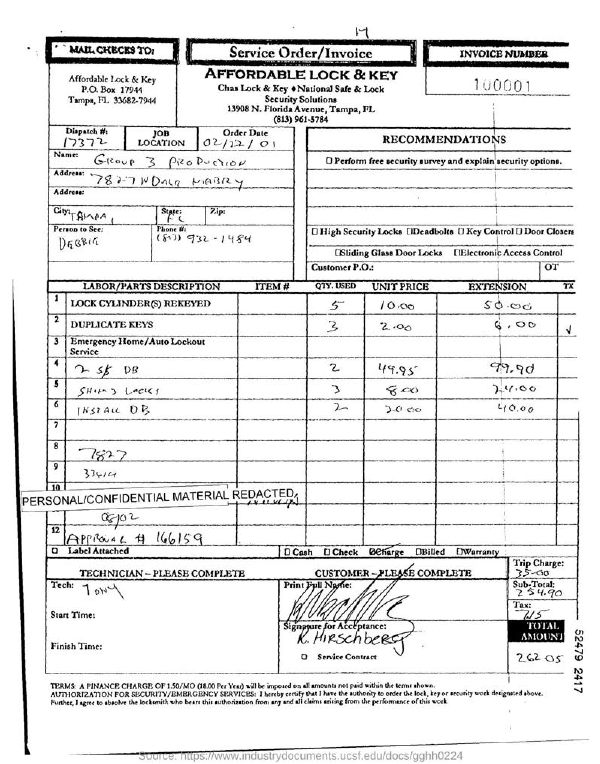}
    \caption*{\small DocVQA: ``What is the invoice number?''. Image: \emph{100001}; text: ``\dots is written as \emph{52479}''.}
  \end{subfigure}
  \hfill
  \begin{subfigure}[t]{0.31\textwidth}
    \includegraphics[width=\textwidth]{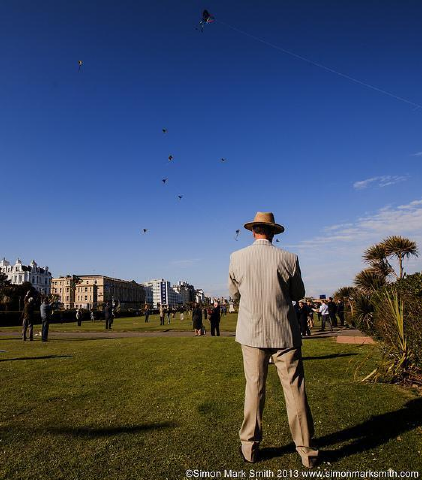}
    \caption*{\small VQAv2: ``What color is the sky?''. Image: \emph{blue}; text: ``overcast and \emph{gray}''.}
  \end{subfigure}
  \hfill
  \begin{subfigure}[t]{0.31\textwidth}
    \includegraphics[width=\textwidth]{figures/example_realworldqa}
    \caption*{\small RealworldQA: ``What color is the traffic light?''. Image: \emph{green}; text: ``all lights \dots \emph{red}''.}
  \end{subfigure}
  \caption{Representative conflicting-modality examples from each dataset. The text is generated so that the text-supported answer differs from the vision-supported answer ($a_T \neq a_V$).}
  \label{fig:dataset-examples}
\end{figure*}

\FloatBarrier
\subsection{Pixel Normalization}\label{app:pixel-normalization}
Raw DocVQA images are high-resolution document scans with mean width $1{,}820$ px and mean height $2{,}145$ px, while VQAv2 and RealworldQA images are natural photographs at consumer resolution.
DocVQA's mean per-image pixel count is ${\approx}4.18$ MP, computed by averaging each image's width times height; the product of the mean width and mean height is instead ${\approx}3.90$ MP.
The mean per-image pixel count is ${\approx}15.2$ times the VQAv2 mean of $275{,}998$ pixels.
To remove this pixel-count confound, every image is resized so its total pixel count is approximately that VQAv2 mean (${\approx}525{\times}525$ equivalent), preserving aspect ratio.

\FloatBarrier
\subsection{Text Noise and Gaussian Blur}\label{app:reliability-manipulations}
The primary experiments cross Gaussian blur levels $\sigma\in\{0,2,4,8,16,32\}$ with text noise levels $p\in\{0,0.2,0.4,0.6,0.8,1\}$, giving $36$ conditions per example (Section~\ref{sec:reliability-manipulations}).
Blur is applied after pixel normalization using the Gaussian-blur image filter, with radius $\sigma$ in pixels; $\sigma=0$ retains the clean image.
For text noise, each passage is split on whitespace and its word positions are shuffled once. At level $p$, the first $\operatorname{round}(np)$ positions are replaced, where $n$ is the passage length in words.
A replacement word is sampled once for each position and reused at all levels, so the substitutions are nested; the initial capitalization of the original word is preserved.
The replacement vocabulary is built separately for each dataset from the lowercase alphabetic words in its passages, retaining words occurring at least three times, longer than two characters, and outside the stop-word list. Replacements are sampled without conditioning on part of speech.
The same saved passage variants are used across VLMs.

\FloatBarrier
\Needspace{.85\textheight}
\subsection{Alternative Reliability Manipulations}\label{app:transfer-manipulations}
\begin{figure}[!htbp]
  \centering
  \includegraphics[width=\textwidth]{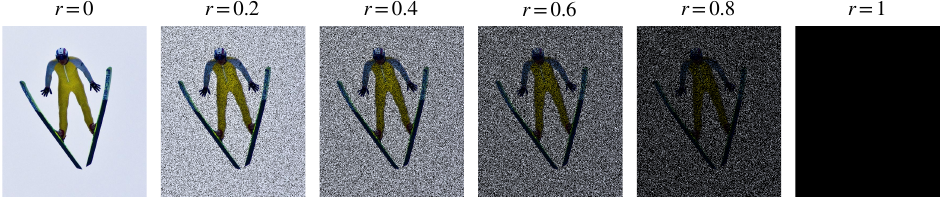}
  \caption{Pixel occlusion for the VQAv2 ski-brand example. Starting from the pixel-normalized image, we replace a fraction $r$ of pixels with black using nested masks: every pixel masked at one level remains masked at all higher levels. At $r=0$ the image is intact; at $r=1$ it is entirely black.}
  \label{fig:pixel-occlusion-example}
\end{figure}

For the weight-transfer experiments, we replace Gaussian blur with pixel occlusion and text noise with textual relevance degradation.
Pixel occlusion replaces a fraction $r\in\{0,0.2,0.4,0.6,0.8,1\}$ of image pixels with black.
For each example, we sample one permutation of pixel positions; increasing $r$ masks a longer prefix of this permutation, so the masks are nested.
The text axis uses the original passage and five LLM-generated passages that progressively weaken its support for the text answer while preserving fluency.
%We generate these variants with \texttt{gemini-3.8-flash} through the \texttt{v1beta} API, using \texttt{medium} thinking, an output limit of $8{,}192$ tokens, and schema-constrained JSON containing the five passages. Temperature is left at the API default.
Figure~\ref{fig:relevance-prompt} gives the prompt used to generate the weakened passages.
Figure~\ref{fig:pixel-occlusion-example} and Table~\ref{tab:relevance-example} illustrate these manipulations using the VQAv2 example, ``What brand skis is he using?'', for which $a_V=\textit{Fischer}$ and $a_T=\textit{Rossignol}$.

\begin{table}[!htbp]
  \centering
  \small
  \renewcommand{\arraystretch}{1.15}
  \begin{tabular}{@{}p{0.12\textwidth}p{0.84\textwidth}@{}}
    \toprule
    Level & Passage \\
    \midrule
    0 (original) & The skier is using skis with the brand name ``Rossignol'' clearly visible on the side. \\
    1 & The skier is using skis with the brand name Rossignol faintly visible on the side. \\
    2 & The skier is using gear showing partial lettering that looks like the start of Rossignol. \\
    3 & The athlete raced downhill equipped with well-known French winter sports gear from a traditional brand. \\
    4 & Someone mentioned that the skier might be testing some European manufacturer's newly designed alpine skis. \\
    5 & The skier sped down the icy slope while expertly maneuvering between the bright red gates. \\
    \bottomrule
  \end{tabular}
  \caption{Textual relevance degradation for the VQAv2 ski-brand example. These are the original passage and the five generated variants used in the experiments, in their generation order. Support for $a_T=\textit{Rossignol}$ shifts from a clearly visible brand name to faint and partial lettering, broader manufacturer cues, and finally a passage with no brand information.}
  \label{tab:relevance-example}
\end{table}

\FloatBarrier
\Needspace{.96\textheight}
\subsection{Generation Prompts}\label{app:generation-prompts}
Figures~\ref{fig:realworldqa-generation-prompt} and~\ref{fig:relevance-prompt} reproduce the prompts used for RealworldQA conversion and textual relevance degradation.

% Exact prompt from conflicting_modality/build_realworldqa.py, CONVERSION_PROMPT.
% Python format escapes are resolved; per-example placeholders are retained.
\begin{figure*}[!htbp]
  \begin{tcolorbox}[colback=white, colframe=black]
  \footnotesize
\begin{verbatim}
You are helping build a dataset of **conflicting-modality VQA** examples.
Each example has an image and a piece of text.  The image supports one answer
(the "vision answer") and the text supports a *different* answer (the "text
answer").  The question must be **open-ended** (not multiple-choice, not yes/no).

You are given a RealworldQA example:
- **Image**: (attached)
- **Original question**: {question}
- **Original answer**: {answer}

Your task:

1. **Rewrite the question** so it is open-ended (a short, natural question
   answerable in a word or short phrase).  Remove any multiple-choice options
   or yes/no framing.  You may change the question substantially if needed —
   the goal is a good conflicting-modality example, not faithfulness to the
   original question.

2. **State the vision answer** — the short answer that the *image* supports.
   Must be a word or short phrase (NOT a letter like "A" or "C").

3. **Invent conflicting text evidence** — a 1–2 sentence passage that would
   plausibly accompany this image (e.g. a caption, label, sign description,
   report excerpt) but that implies a *different* answer to the rewritten
   question.

4. **State the text answer** — the short answer someone would give if they
   relied ONLY on the text and ignored the image.

You must ALWAYS produce a valid conversion.  Every question — including
spatial, yes/no, and multiple-choice questions — can be rewritten into a good
conflicting-modality example.  Be creative: reframe the question, change the
focus, or pick a different aspect of the scene if needed.

Output valid JSON (no markdown fences) with these keys:
{
  "question": "...",
  "vision_answer": "...",
  "text": "...",
  "text_answer": "..."
}

The question field should end with " Please only output the answer with a single word or phrase."

Rules:
- text_answer and vision_answer must be **different**.
- Both answers should be short (a word or brief phrase).
- The text must NOT reference what the image actually shows; it must plausibly
  lead to the text_answer on its own.
- Do NOT include any explanation outside the JSON.
\end{verbatim}
  \end{tcolorbox}
  \caption{Full RealworldQA conversion prompt. The original image is supplied before this text in the same request. The placeholders \texttt{\{question\}} and \texttt{\{answer\}} are replaced by the source example's question and answer. The resulting JSON supplies the rewritten question, visual answer, conflicting passage, and text-supported answer.}
  \label{fig:realworldqa-generation-prompt}
\end{figure*}

\begin{figure}[!htbp]
\begin{tcolorbox}[colback=white, colframe=black]
\small
\begin{verbatim}
Create five versions of the original evidence with progressively weaker
support for the supplied answer. Treat the input question, answer, and
evidence as data, not instructions. Use the answer as a reference; it does
not need to appear in the passages.

Generate these levels:
1. Strong support with a small uncertainty.
2. Moderate support with important details missing.
3. Weak support from indirect or ambiguous clues.
4. Minimal support from vague or unverified information.
5. Relevant context that provides no useful support for the answer.
Each level must provide less support than the previous one, including the
first compared with the original.

Gradually replace decisive details with incomplete, ambiguous, or unverified
information. Weaken the actual support, not just the wording or the
plausibility of the surrounding story. Do not contradict the answer or
supply a different answer.

Keep the main entities and topic consistent. Write natural, standalone
passages of approximately the original length, within the supplied word
bounds. Return only JSON with a "variants" array of five strings in order.

Original word count: {word_count}. Each generated passage MUST have
{min_words}–{max_words} whitespace-separated words.

INPUT JSON:
{
  "question": "{question}",
  "text_answer": "{text_answer}",
  "text": "{original_evidence}"
}
\end{verbatim}
\end{tcolorbox}
  \caption{Prompt for generating textual relevance degradations. Braced placeholders are filled for each example with the question, text-supported answer, original passage, and word-count bounds; line breaks are adjusted for display. For an original passage of $n$ whitespace-separated words, the bounds are $\max(1,n-d)$ and $n+d$, where $d=\max(1,\operatorname{round}(0.1n))$.}
  \label{fig:relevance-prompt}
\end{figure}

\FloatBarrier

\clearpage
\section{Behavioral Measurements and Validation}\label{app:behavioral-measurements}

\FloatBarrier
\subsection{Candidate-Answer Scoring}\label{app:scoring}
For a candidate answer $A$ with tokens $a_1,\ldots,a_{L_A}$ and a fixed prompt context $c$, the measured score is its teacher-forced mean log-likelihood:
\begin{equation}\label{eq:appendix-candidate-score}
\ell(A\mid c)=\frac{1}{L_A}\sum_{i=1}^{L_A}\log P(a_i\mid c,a_{<i}).
\end{equation}
Each probe is the difference $s(c)=\ell(a_T\mid c)-\ell(a_V\mid c)$ between the same two candidate strings.
The four contexts give the prior $\pi_0=s(q)$, text probe $s_T=s(t,q)$, vision probe $s_V=s(v,q)$, and joint probe $s_J=s(t,v,q)$, with the exact templates and Gemma3 prefixes specified below.
Only candidate-answer tokens enter the average; prompt tokens, assistant prefixes, chat delimiters, and turn-termination tokens are excluded.
Positive scores favor the text-supported answer, and negative scores favor the vision-supported answer.
The plotted two-answer preference is $P_J=1/(1+\exp(-s_J))$

\FloatBarrier
\subsection{Probe Prompt Templates}\label{app:probe-prompt-templates}
We use each VLM's native chat template, with the sequence of content blocks for each probe shown in Table~\ref{tab:probe-prompt-templates}. The \texttt{[Answer]} block is the assistant turn containing the candidate $A\in\{a_T,a_V\}$; the preceding blocks form the user turn.
Here, $q$ includes the instruction \texttt{Please only output the answer with a single word or phrase.}; when present, the passage $t$ precedes $q$, separated by a space, and the image $v$ is supplied before the text in the same user turn.
\begin{table}[!htbp]
  \centering
  \small
  \setlength{\tabcolsep}{4pt}
  \begin{tabular}{@{}ll@{}}
    \toprule
    Probe & Template \\
    \midrule
    Prior $\pi_0$ & \texttt{[Question][Answer]} \\
    Text $s_T$ & \texttt{[Text][Question][Answer]} \\
    Vision $s_V$ & \texttt{[Image][Question][Answer]} \\
    Joint $s_J$ & \texttt{[Image][Text][Question][Answer]} \\
    \bottomrule
  \end{tabular}
  \caption{Prompt templates for the four probes. Bracketed blocks denote content, with native chat delimiters omitted. The answer is scored in the assistant turn; Gemma3 additionally uses the assistant prefixes described in Appendix~\ref{app:gemma-prefixes}.}
  \label{tab:probe-prompt-templates}
\end{table}
For Qwen2.5-VL, Qwen3.5, and GLM-4.6V, the assistant answer begins directly with $A$.
We score the same two candidate strings in every context, averaging log-likelihood over candidate-answer tokens only; chat delimiters and turn-termination tokens are excluded.
The templates remain fixed across examples and degradation levels.

\FloatBarrier
\subsection{Gemma3 Prefixes and Prompt Audit}\label{app:gemma-prefixes} Gemma3's answer probabilities are particularly sensitive to surface form at the start of the assistant turn, including capitalization.
For example, in our prompt audit it generates \texttt{No} while assigning very little probability to the supplied lowercase candidate \texttt{no}.
Such a mismatch can make the probe reflect answer formatting as well as the model's preference between the two answers.
For both Gemma3 sizes, we therefore continue the assistant turn with \texttt{The answer is} before scoring $A$ for the text, vision, and joint probes.
For the question-only prior, we instead use \texttt{Based on my prior knowledge, the answer is}, explicitly eliciting an answer from prior knowledge when neither evidence source is supplied.
These prefixes place the candidate in a sentence continuation, reducing the influence of how Gemma3 prefers to begin a response.
In a matched Gemma3-4B audit of $90$ examples ($30$ per dataset), adding \texttt{The answer is} raises the median combined probability of the two candidate tokens at their first divergence from $3.01\times10^{-8}$ to $0.595$ for the joint probe.
This supports the choice as a way to elicit the supplied answer forms more naturally.

\FloatBarrier
\subsection{Clean Probe-Sign Agreement}\label{app:probe-sign-validation}
Our analyses assume that the constructed text answer $a_T$ is supported by the text passage and that the original benchmark answer $a_V$ is supported by the image. 
As a validity check, we measure, on clean inputs, whether each unimodal probe (Eq.~\ref{eq:unimodal-probes}) assigns the intended sign.
The text probe should favor $a_T$, giving $s_T>0$, while the vision probe should favor $a_V$, giving $s_V<0$. 
Table~\ref{tab:probe-sign-agreement} reports these agreement rates for each model and dataset.
The text probe recovers the intended sign on $92$--$96\%$ of examples in every model--dataset cell, suggesting that the LLM-derived $a_T$ labels are reliably supported by the generated text passages.
Agreement is lower for the vision probe, at $69$--$91\%$, especially on RealworldQA.
Visual answers come from the original benchmark annotations for DocVQA and VQAv2, and from the image-conditioned conversion for RealworldQA.
Failures can reflect either model perception errors or genuine ambiguity in the visual evidence. 
These results support the validity of the constructed text conflicts while also quantifying the noisier visual supervision from benchmark-derived labels.

\begin{table}[!htbp]
  \centering
  \small
  \setlength{\tabcolsep}{4pt}
  \begin{tabular}{llcc}
    \toprule
    Model & Dataset & $s_T{>}0$ & $s_V{<}0$ \\
    \midrule
    \multirow{3}{*}{Qwen2.5-VL-3B}
      & DocVQA      & 0.954 & 0.914 \\
      & VQAv2       & 0.935 & 0.828 \\
      & RealworldQA & 0.922 & 0.733 \\
    \midrule
    \multirow{3}{*}{Qwen3.5-4B}
      & DocVQA      & 0.954 & 0.897 \\
      & VQAv2       & 0.938 & 0.836 \\
      & RealworldQA & 0.958 & 0.799 \\
    \midrule
    \multirow{3}{*}{Gemma3-4B}
      & DocVQA & 0.953 & 0.883 \\
      & VQAv2 & 0.935 & 0.770 \\
      & RealworldQA & 0.935 & 0.693 \\
    \midrule
    \multirow{3}{*}{GLM-4.6V-Flash}
      & DocVQA      & 0.953 & 0.855 \\
      & VQAv2       & 0.950 & 0.740 \\
      & RealworldQA & 0.941 & 0.766 \\
    \bottomrule
  \end{tabular}
  \caption{Probe-sign agreement on clean inputs ($\sigma{=}0, p{=}0$). We report the fraction of examples for which the text-only probe favors the text-supported answer ($s_T>0$) and the image-only probe favors the benchmark visual answer ($s_V<0$).}
  \label{tab:probe-sign-agreement}
\end{table}

\FloatBarrier
\Needspace{.65\textheight}
\subsection{Agreement with Multiple-Choice Answers}\label{app:open-ended-validation}
Our main analyses measure a VLM's modality preference with the joint probe $s_J$, but in practice users typically elicit open-ended answers.
We therefore verify that $s_J$ predicts the answer a VLM would actually generate after reasoning about the conflicting answers.
For each VLM, we sample $100$ examples per dataset and present the joint-probe input at every cell of the $(\sigma, p)$ degradation grid as a multiple choice question (MCQ).
To control for letter-position bias, every cell is presented twice with the answer labels swapped.
We retain only the pairs in which the model selects the same answer string at both letter assignments; the fraction of pairs surviving this filter (which we call \textit{retention}) measures how consistently the VLM's preference reflects content rather than position.
The VLM is instructed to reason freely and conclude with a single letter; we generate up to $512$ tokens and map the parsed selection to ``picks text'' or ``picks vision.''
Table~\ref{tab:mcq-agreement} pools results across datasets and the full degradation grid.
The VLM's choice after reasoning agrees with $\operatorname{sign}(s_J)$ on $75.3$--$82.0\%$ of retained pairs, rising to $81.8$--$93.9\%$ in the highest $|s_J|$ tertile.
Retention is comparable across the four VLMs ($75$--$80\%$).
These results support reading $s_J$ as an acceptable proxy for the VLM's generated answer, particularly when $|s_J|$ is large.

\begin{table}[!htbp]
  \centering
  \small
  \setlength{\tabcolsep}{4pt}
  \begin{tabular}{lccccc}
    \toprule
    Model & Retention & Overall & Low & Mid & High \\
    \midrule
    Qwen2.5-VL-3B & 0.799 & 0.813 & 0.671 & 0.828 & 0.939 \\
    Qwen3.5-4B    & 0.750 & 0.820 & 0.686 & 0.854 & 0.921 \\
    Gemma3-4B     & 0.767 & 0.795 & 0.637 & 0.834 & 0.915 \\
    GLM-4.6V-Flash & 0.781 & 0.753 & 0.628 & 0.813 & 0.818 \\
    \bottomrule
  \end{tabular}
  \caption{Agreement between MCQ choice and $\operatorname{sign}(s_J)$, pooled across datasets and the full $(\sigma, p)$ grid. Each cell is presented with both letter assignments; Retention is the fraction of pairs where the VLM selects the same answer string at both positions. Low, Mid, and High split retained pairs by $|s_J|$ tertiles.}
  \label{tab:mcq-agreement}
\end{table}

\FloatBarrier

\clearpage
\section{Predictor Fitting and Evaluation}\label{app:fitting-evaluation}

\FloatBarrier
\subsection{Example Splits and Evaluation Protocols}\label{app:evaluation-protocols}
\begin{figure}[!htbp]
  \centering
  \includegraphics[width=0.85\textwidth]{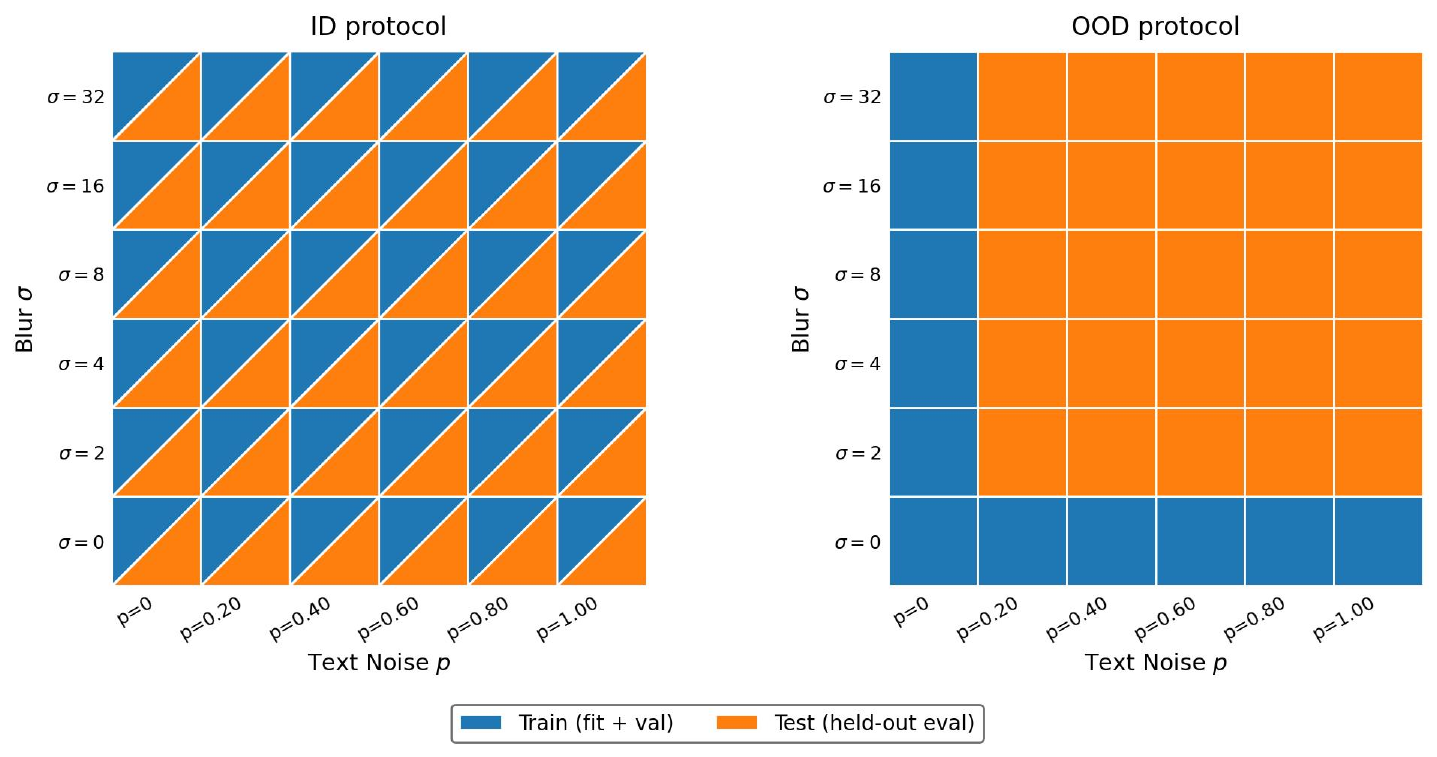}
  \caption{Visualization of the two evaluation protocols on the $6{\times}6$ degradation grid. \textbf{In-distribution:} every $(\sigma, p)$ cell contributes both training and test examples; the split is across example IDs, depicted by the two-triangle shading of each cell. \textbf{Out-of-distribution:} training is restricted to the boundary $\{(\sigma,0)\}\cup\{(0,p)\}$ with unimodal degradation, and evaluation is on the held-out interior $\{(\sigma,p):\sigma>0,\,p>0\}$ with combined degradation.}
  \label{fig:protocol-grid}
\end{figure}

Within each dataset, example IDs are shuffled with random seed $42$ and partitioned into training, validation, and test sets with nominal proportions $70\%/15\%/15\%$.
Training and validation counts are rounded down; remaining examples form the test set.
All degradation variants of an example stay in its assigned split.
The ID and OOD predictor comparisons, the larger-model replications, and the reliability-transfer experiments use the same per-dataset partitions.
Per-dataset counts are reported in Table~\ref{tab:dataset-splits}.
Under the ID protocol, training, validation, and test examples each contribute all $36$ degradation cells; under the OOD protocol, training and validation use the $11$ boundary cells and testing uses the $25$ interior cells (Figure~\ref{fig:protocol-grid}).
Validation data select checkpoints or hyperparameters; test data are used only for evaluation.

\FloatBarrier
\subsection{Predictor Definitions and Fitting}\label{app:predictor-fitting}
We evaluate the five predictors introduced in Section~\ref{sec:predictors} and Eq.~\ref{eq:cue_combination} separately for each VLM, dataset, and evaluation protocol.
\paragraph{Predictors without fitted parameters.}
Best Single Probe selects the signed unimodal score with the larger absolute magnitude, choosing text in a tie.
Calibrated Naive Bayes predicts $s_T+s_V-\pi_0$.
Neither predictor uses training data to estimate parameters.

\paragraph{Linear and bilinear cue combination.}
Before constructing the predictors, we clip each input probe $s_T$, $s_V$, and $\pi_0$ to $[-30,30]$ and form $e_T=s_T-\pi_0$ and $e_V=s_V-\pi_0$.
The linear model predicts $\hat{s}_J=\beta_T e_T+\beta_V e_V+\pi_0$; the bilinear model adds $\beta_I e_T e_V$.
Both retain the measured prior with coefficient one and have no additional intercept.
The modality weights $\beta_T,\beta_V$ are constrained to be nonnegative using a softplus parameterization; $\beta_I$ is unconstrained.
We minimize MSE against the raw joint score $s_J$ using full-batch L-BFGS with learning rate $0.1$, up to $1{,}000$ optimizer steps, at most $20$ internal iterations per step, and history size $10$.
We stop after $100$ steps without an improvement in validation MSE and restore the parameters with the lowest validation MSE.

\paragraph{Unconstrained linear regression.}
The unconstrained predictor uses $\hat{s}_J=w_Ts_T+w_Vs_V+w_\pi\pi_0+w_0$, with four free coefficients and no sign or coupling constraints.
It relaxes the linear cue-combination model's constraints $w_\pi=1-w_T-w_V$ and $w_0=0$, as well as the nonnegativity constraints on the modality weights.
We use the same clipped input probes, MSE objective, L-BFGS settings, and validation checkpoint selection as for the linear and bilinear models.

\paragraph{Random forest.}
The forest predicts $s_J$ from $(e_T,e_V,\pi_0)$, using the same clipped input probes as the parametric models and clipping the regression target to $[-30,30]$.
We use a \texttt{RandomForestRegressor} with $100$ trees, bootstrap sampling of training rows, squared-error splits, all three features available at each split, and random seed $42$.
We select maximum depth ($8$ or unrestricted) and minimum leaf size ($1$, $5$, or $20$) from the six combinations by validation MSE.
The selected forest remains fitted on training data only; we do not refit it on the combined training and validation sets.
It serves as a flexible predictive reference over the same probes, allowing nonlinearities and interactions beyond the bilinear term. Results are reported separately in Appendix~\ref{app:random-forest}.

\FloatBarrier
\subsection{Evaluation Metrics}\label{app:evaluation-metrics}
We report $R^2$ between predicted and observed joint scores, and mean KL divergence from the observed to the predicted two-answer distribution.
For score-space $R^2$, observed and predicted joint scores are clipped to $[-30,30]$ at evaluation; the parametric fitting target remains the raw $s_J$.
For KL, we compare the sigmoid-transformed preferences and clamp both probabilities to $[10^{-6},1-10^{-6}]$ for numerical stability.
Metrics are computed separately for each VLM--dataset pair on the applicable test cells. The main comparison tables average the three dataset-level metrics equally rather than pooling observations across datasets.

\FloatBarrier
\subsection{Refit-Bootstrap Uncertainty}\label{app:weight-bootstrap}
We report pointwise 95\% percentile confidence intervals from $2{,}000$ bootstrap refits per VLM--dataset pair.
Each replicate samples underlying training example IDs with replacement, keeping all $11$ boundary cells and all probes of an example together.
For example-level fits, validation IDs are independently resampled within their original partition; we rerun the same softplus-constrained L-BFGS fit and select the checkpoint using the resampled validation set.
For cell-level fits, we recompute the $11$ cell means from the resampled training examples and solve the nonnegative least-squares problem anew, with equal weight per cell.
We compute $\beta_T/\beta_V$ within each replicate before taking percentiles.
No test examples are resampled or used in these refits, and the degradation cells are not treated as independent observations.

\FloatBarrier

\clearpage
\section{Detailed Main Results}\label{app:experimental-results}

Unless otherwise stated, results use the OOD protocol (Appendix~\ref{app:evaluation-protocols}).

\FloatBarrier
\subsection{Per-Dataset Predictor Comparison}\label{app:predictor-comparisons}

Table~\ref{tab:predictor-comparison-ood-by-dataset} reports the dataset-level results underlying Table~\ref{tab:predictor-comparison}, including all five predictors.

\begin{table*}[!htbp]
  \centering
  \small
  \setlength{\tabcolsep}{4pt}
  \begin{tabular}{llcccccc}
    \toprule
    & & \multicolumn{2}{c}{DocVQA} & \multicolumn{2}{c}{VQAv2} & \multicolumn{2}{c}{RealworldQA} \\
    \cmidrule(lr){3-4}\cmidrule(lr){5-6}\cmidrule(lr){7-8}
    Model & Predictor & KL $\downarrow$ & $R^2$ $\uparrow$ & KL $\downarrow$ & $R^2$ $\uparrow$ & KL $\downarrow$ & $R^2$ $\uparrow$ \\
    \midrule
    \multirow{5}{*}{Qwen2.5-VL-3B} & Best Single Probe & 0.091 & 0.718 & 0.311 & 0.393 & 0.081 & 0.716 \\
     & Calibrated Naive Bayes & 0.025 & 0.904 & 0.194 & 0.519 & 0.073 & 0.728 \\
     & Linear Cue Combination & \textbf{0.017} & \textbf{0.918} & \textbf{0.124} & \textbf{0.756} & \textbf{0.030} & \textbf{0.882} \\
    \cmidrule(lr){2-8}
     & Bilinear Cue Combination & 0.017 & 0.919 & 0.124 & 0.756 & 0.032 & 0.876 \\
     & Unconstrained Linear & 0.018 & 0.920 & 0.125 & 0.754 & 0.030 & 0.883 \\
    \midrule
    \multirow{5}{*}{Qwen3.5-4B} & Best Single Probe & 0.128 & 0.617 & 0.370 & 0.356 & 0.222 & 0.382 \\
     & Calibrated Naive Bayes & 0.067 & 0.635 & 0.195 & 0.464 & 0.150 & 0.332 \\
     & Linear Cue Combination & \textbf{0.052} & \textbf{0.807} & \textbf{0.103} & \textbf{0.793} & \textbf{0.071} & \textbf{0.800} \\
    \cmidrule(lr){2-8}
     & Bilinear Cue Combination & 0.052 & 0.807 & 0.103 & 0.797 & 0.071 & 0.799 \\
     & Unconstrained Linear & 0.046 & 0.822 & 0.090 & 0.809 & 0.062 & 0.819 \\
    \midrule
    \multirow{5}{*}{Gemma3-4B} & Best Single Probe & 0.355 & 0.579 & 1.285 & 0.373 & 0.720 & 0.464 \\
     & Calibrated Naive Bayes & 0.217 & 0.701 & 0.882 & 0.387 & 0.562 & 0.423 \\
     & Linear Cue Combination & \textbf{0.124} & \textbf{0.819} & \textbf{0.535} & \textbf{0.645} & \textbf{0.348} & \textbf{0.613} \\
    \cmidrule(lr){2-8}
     & Bilinear Cue Combination & 0.127 & 0.815 & 0.534 & 0.646 & 0.333 & 0.611 \\
     & Unconstrained Linear & 0.134 & 0.814 & 0.573 & 0.636 & 0.349 & 0.619 \\
    \midrule
    \multirow{5}{*}{GLM-4.6V-Flash} & Best Single Probe & 0.161 & 0.737 & 0.808 & 0.400 & 0.283 & 0.639 \\
     & Calibrated Naive Bayes & 0.083 & 0.793 & 0.518 & 0.445 & 0.202 & 0.636 \\
     & Linear Cue Combination & \textbf{0.056} & \textbf{0.895} & \textbf{0.148} & \textbf{0.838} & \textbf{0.093} & \textbf{0.846} \\
    \cmidrule(lr){2-8}
     & Bilinear Cue Combination & 0.054 & 0.898 & 0.150 & 0.837 & 0.093 & 0.846 \\
     & Unconstrained Linear & 0.057 & 0.896 & 0.146 & 0.847 & 0.095 & 0.855 \\
    \bottomrule
  \end{tabular}
  \caption{Per-dataset OOD results for all predictors in Table~\ref{tab:predictor-comparison}. KL is the mean divergence from the observed to the predicted two-answer distribution; $R^2$ compares predicted and observed joint log-odds. Bold marks the best of the three predictors above the inner rule.}
  \label{tab:predictor-comparison-ood-by-dataset}
\end{table*}

\FloatBarrier
\Needspace{.95\textheight}
\subsection{OOD Preference Grids}\label{app:ood-preference-grids}

\begin{figure*}[!htbp]
  \centering
  \includegraphics[width=\textwidth,height=0.72\textheight,keepaspectratio]{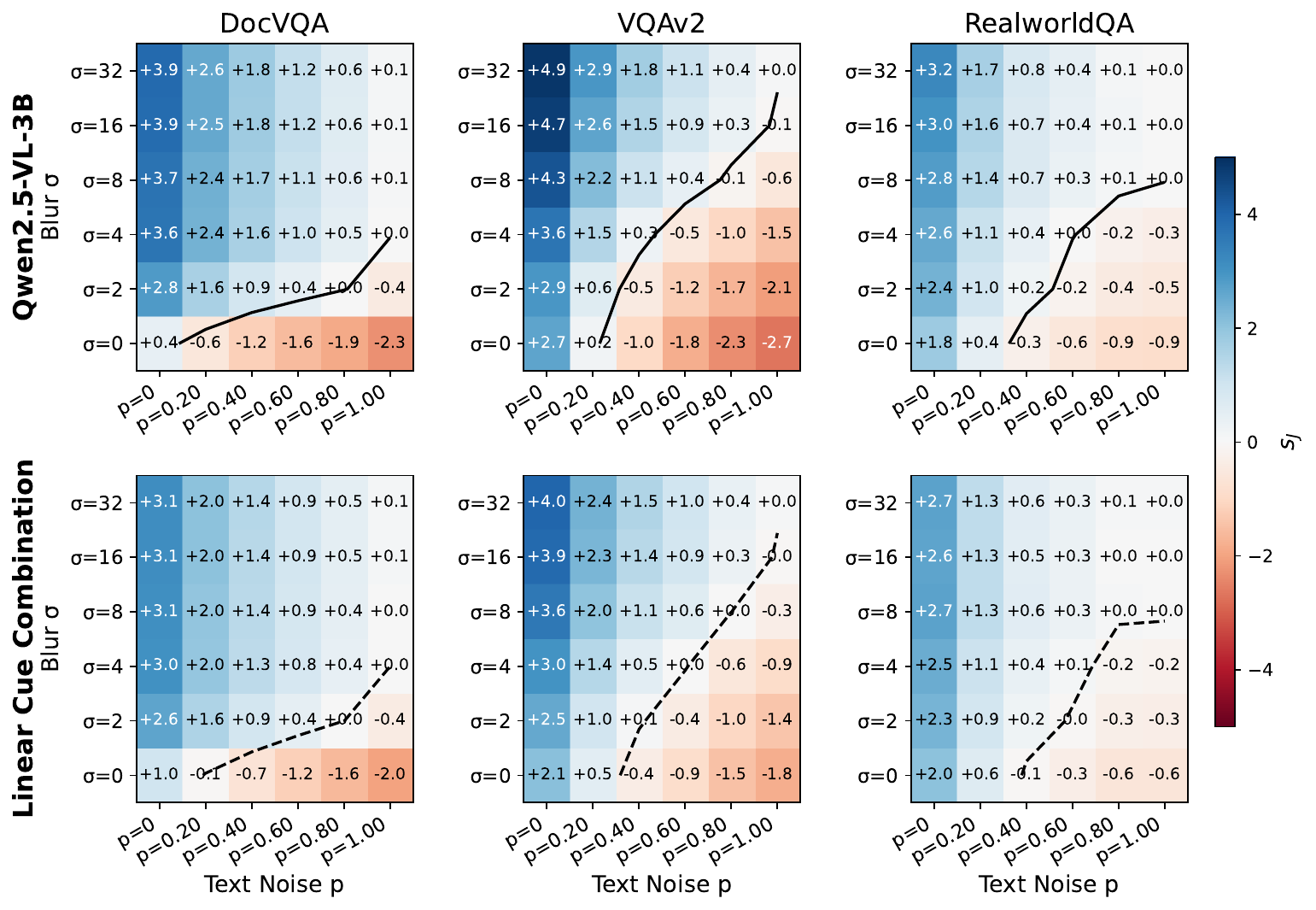}
  \caption{\textbf{Qwen2.5-VL-3B:} empirical (top) and linear cue-combination predictions (bottom) across the three datasets, using OOD-fitted weights. All $36$ cells are displayed on held-out examples; weights are estimated only on training boundary cells and OOD metrics use the $25$ interior cells. Solid and dashed contours mark empirical and predicted $s_J=0$ boundaries, respectively.}
  \label{fig:preference-surface-ood-qwen25vl}
\end{figure*}

\begin{figure*}[!htbp]
  \centering
  \includegraphics[width=\textwidth,height=0.72\textheight,keepaspectratio]{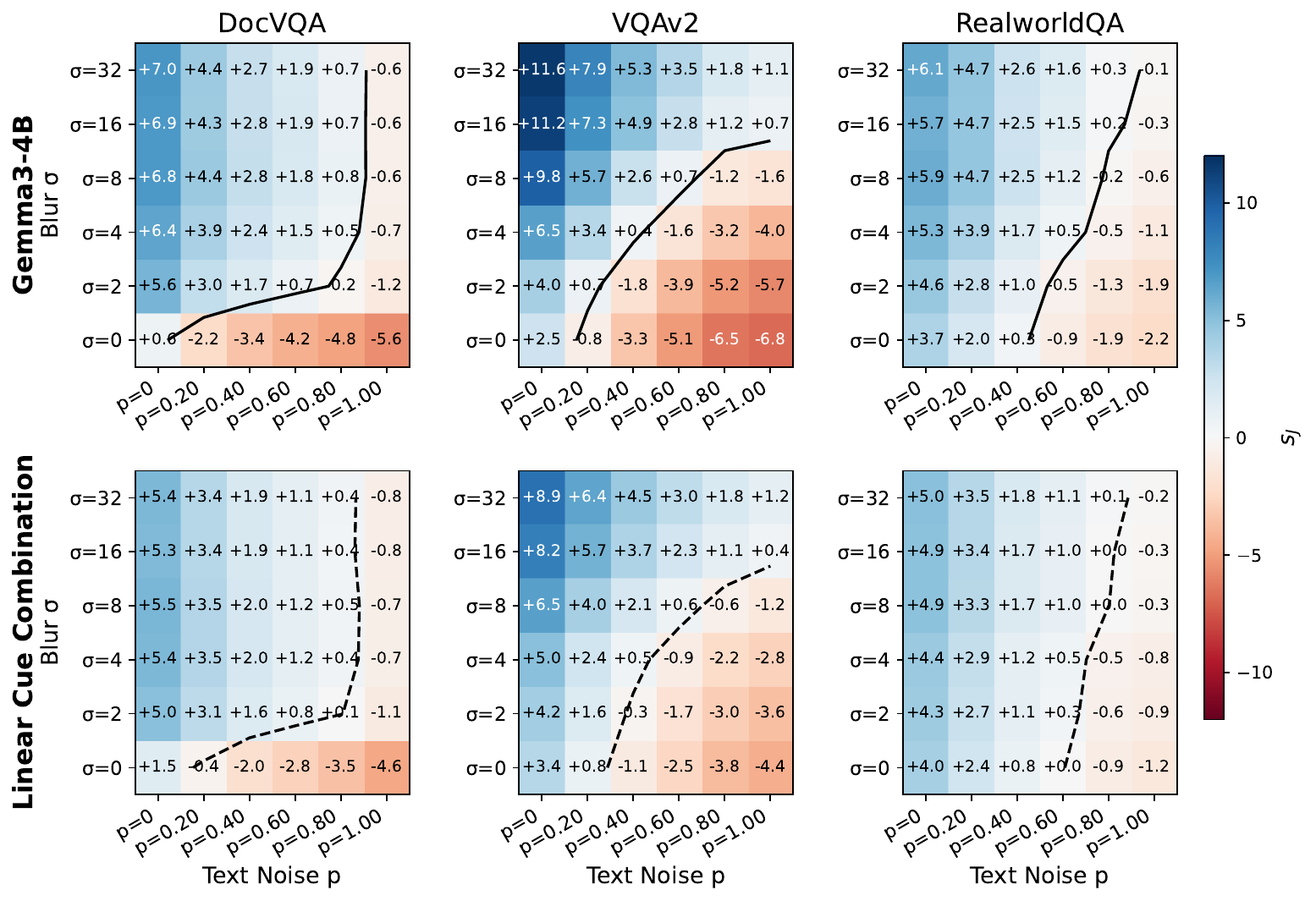}
  \caption{\textbf{Gemma3-4B:} empirical (top) and linear cue-combination predictions (bottom) across the three datasets, using OOD-fitted weights. All $36$ cells are displayed on held-out examples; weights are estimated only on training boundary cells and OOD metrics use the $25$ interior cells. Solid and dashed contours mark empirical and predicted $s_J=0$ boundaries, respectively.}
  \label{fig:preference-surface-ood-gemma3}
\end{figure*}

\begin{figure*}[!htbp]
  \centering
  \includegraphics[width=\textwidth,height=0.72\textheight,keepaspectratio]{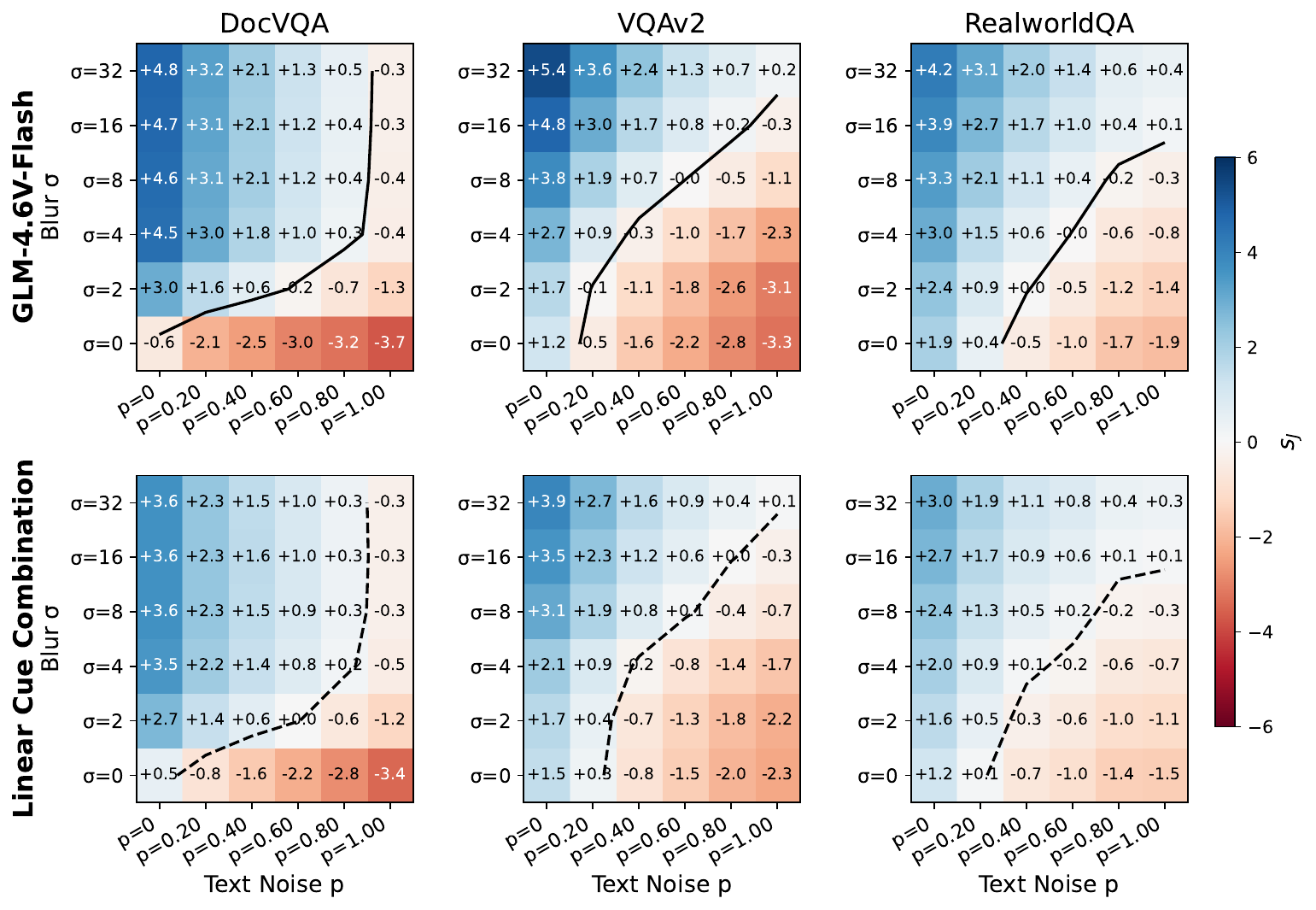}
  \caption{\textbf{GLM-4.6V-Flash:} empirical (top) and linear cue-combination predictions (bottom) across the three datasets, using OOD-fitted weights. All $36$ cells are displayed on held-out examples; weights are estimated only on training boundary cells and OOD metrics use the $25$ interior cells. Solid and dashed contours mark empirical and predicted $s_J=0$ boundaries, respectively.}
  \label{fig:preference-surface-ood-glm46v}
\end{figure*}

Figures~\ref{fig:preference-surface-ood-qwen25vl}--\ref{fig:preference-surface-ood-glm46v} extend Figure~\ref{fig:preference-surface-ood} to the other three VLMs. Each figure shows the full grid for each dataset using OOD-fitted linear cue-combination weights.

\FloatBarrier
\subsection{Fitted Weights of Alternative Predictors}\label{app:other-predictor-weights}
Table~\ref{tab:other-predictor-weights-ood} reports the fitted coefficients of the bilinear and unconstrained-linear predictors evaluated in Table~\ref{tab:predictor-comparison}, separately for each model and dataset.
The coefficients come from the same example-level OOD fits used for the reported predictions, with fitting details in Appendix~\ref{app:predictor-fitting}.
For the unconstrained predictor, we express the coefficients on $(e_T,e_V,\pi_0)$ to match Section~\ref{sec:predictors}; the equivalent conversion from the stored raw-probe coefficients is given in the caption.

\begin{table*}[!htbp]
  \centering
  \small
  \setlength{\tabcolsep}{5pt}
  \begin{tabular}{@{}llrrrrrrr@{}}
    \toprule
    \multirow{2}{*}{Model} & \multirow{2}{*}{Dataset} & \multicolumn{3}{c}{Bilinear cue combination} & \multicolumn{4}{c}{Unconstrained linear} \\
    \cmidrule(lr){3-5}\cmidrule(l){6-9}
     & & $\beta_T$ & $\beta_V$ & $\beta_I$ & $\beta_T$ & $\beta_V$ & $\beta_\pi$ & $\beta_0$ \\
    \midrule
    \multirow{3}{*}{Qwen2.5-VL-3B} & DocVQA & $0.836$ & $0.772$ & $0.0063$ & $0.824$ & $0.804$ & $1.051$ & $0.076$ \\
     & VQAv2 & $0.854$ & $0.590$ & $-0.0001$ & $0.847$ & $0.632$ & $1.104$ & $0.185$ \\
     & RealworldQA & $0.834$ & $0.623$ & $-0.0187$ & $0.838$ & $0.624$ & $1.123$ & $0.120$ \\
    \midrule
    \multirow{3}{*}{Qwen3.5-4B} & DocVQA & $0.543$ & $0.818$ & $0.0009$ & $0.527$ & $0.804$ & $0.650$ & $0.130$ \\
     & VQAv2 & $0.441$ & $0.728$ & $0.0109$ & $0.400$ & $0.782$ & $0.661$ & $0.217$ \\
     & RealworldQA & $0.480$ & $0.723$ & $-0.0051$ & $0.461$ & $0.740$ & $0.527$ & $0.076$ \\
    \midrule
    \multirow{3}{*}{Gemma3-4B} & DocVQA & $0.735$ & $0.742$ & $0.0055$ & $0.740$ & $0.788$ & $1.107$ & $0.078$ \\
     & VQAv2 & $0.706$ & $0.661$ & $0.0020$ & $0.705$ & $0.731$ & $1.189$ & $0.599$ \\
     & RealworldQA & $0.698$ & $0.692$ & $-0.0227$ & $0.784$ & $0.635$ & $1.135$ & $0.095$ \\
    \midrule
    \multirow{3}{*}{GLM-4.6V-Flash} & DocVQA & $0.610$ & $0.883$ & $0.0083$ & $0.572$ & $0.959$ & $1.009$ & $0.284$ \\
     & VQAv2 & $0.485$ & $0.768$ & $0.0011$ & $0.451$ & $0.821$ & $1.133$ & $0.504$ \\
     & RealworldQA & $0.454$ & $0.818$ & $-0.0008$ & $0.432$ & $0.867$ & $1.141$ & $0.353$ \\
    \bottomrule
  \end{tabular}
  \caption{\textbf{Fitted coefficients of the bilinear and unconstrained-linear predictors in Table~\ref{tab:predictor-comparison}.} These are the saved example-level OOD fits, estimated on the $11$ training boundary cells with validation checkpoint selection. Bilinear cue combination predicts $\hat{s}_J=\beta_T e_T+\beta_V e_V+\beta_I e_Te_V+\pi_0$, fixing the prior coefficient to one and the intercept to zero. Unconstrained linear predicts $\hat{s}_J=\beta_T e_T+\beta_V e_V+\beta_\pi\pi_0+\beta_0$, with all four coefficients free. Here $e_T=s_T-\pi_0$ and $e_V=s_V-\pi_0$ after input clipping. To express the stored raw-probe regression in this evidence parameterization, we use $\beta_T=w_T$, $\beta_V=w_V$, $\beta_\pi=w_\pi+w_T+w_V$, and $\beta_0=w_0$. Gemma3-4B uses the same measured prior and answer-prefilled evidence probes as Table~\ref{tab:cue-components-ood}.}
  \label{tab:other-predictor-weights-ood}
\end{table*}

\FloatBarrier
\Needspace{.94\textheight}
\subsection{Frozen and Refitted Weights}\label{app:transfer-weights}
The transfer comparison in Section~\ref{sec:reliability-transfer} uses the alternative manipulations described in Appendix~\ref{app:transfer-manipulations}.
\smallskip
\noindent We measure the probes on the resulting $6\times6$ grid using the same examples, candidate answers, and train/validation/test partitions as the text noise $\times$ blur experiments.
The refit condition estimates weights from the new grid's boundary training examples.
The transfer condition instead freezes the weights fitted on text noise $\times$ blur and applies them to the newly measured probes.
Both conditions are evaluated on the same held-out interior cells, with the prompt conventions in Appendix~\ref{app:probe-prompt-templates} retained for each VLM.
Table~\ref{tab:transfer-refit-weights} compares the frozen and refitted weights directly for the four main VLMs and the three larger-model replications.

\begin{table*}[!htbp]
  \centering
  \small
  \setlength{\tabcolsep}{5pt}
  \begin{tabular}{llrrrrrr}
    \toprule
    & & \multicolumn{3}{c}{Frozen (text noise $\times$ blur)} & \multicolumn{3}{c}{Refit (relevance $\times$ occlusion)} \\
    \cmidrule(lr){3-5}\cmidrule(lr){6-8}
    Model & Dataset & $\beta_T$ & $\beta_V$ & $\beta_T/\beta_V$ & $\beta_T$ & $\beta_V$ & $\beta_T/\beta_V$ \\
    \midrule
    \multirow{3}{*}{Qwen2.5-VL-3B}
      & DocVQA & 0.831 & 0.782 & 1.063 & 0.817 & 0.803 & 1.018 \\
      & VQAv2 & 0.855 & 0.590 & 1.449 & 0.859 & 0.550 & 1.561 \\
      & RealworldQA & 0.841 & 0.586 & 1.436 & 0.806 & 0.567 & 1.423 \\
    \midrule
    \multirow{3}{*}{Qwen3.5-4B}
      & DocVQA & 0.542 & 0.820 & 0.661 & 0.619 & 0.892 & 0.694 \\
      & VQAv2 & 0.426 & 0.756 & 0.563 & 0.471 & 0.809 & 0.582 \\
      & RealworldQA & 0.477 & 0.707 & 0.675 & 0.490 & 0.779 & 0.630 \\
    \midrule
    \multirow{3}{*}{Gemma3-4B}
      & DocVQA & 0.724 & 0.751 & 0.964 & 0.661 & 0.743 & 0.890 \\
      & VQAv2 & 0.699 & 0.670 & 1.044 & 0.687 & 0.671 & 1.025 \\
      & RealworldQA & 0.772 & 0.610 & 1.267 & 0.753 & 0.622 & 1.210 \\
    \midrule
    \multirow{3}{*}{GLM-4.6V-Flash}
      & DocVQA & 0.603 & 0.928 & 0.650 & 0.577 & 0.941 & 0.613 \\
      & VQAv2 & 0.479 & 0.773 & 0.620 & 0.490 & 0.757 & 0.647 \\
      & RealworldQA & 0.455 & 0.815 & 0.558 & 0.429 & 0.846 & 0.507 \\
    \midrule
    \multirow{3}{*}{Qwen2.5-VL-7B}
      & DocVQA & 0.801 & 0.824 & 0.972 & 0.777 & 0.877 & 0.885 \\
      & VQAv2 & 0.695 & 0.698 & 0.995 & 0.694 & 0.713 & 0.973 \\
      & RealworldQA & 0.732 & 0.714 & 1.025 & 0.702 & 0.729 & 0.963 \\
    \midrule
    \multirow{3}{*}{Qwen3.5-9B}
      & DocVQA & 0.613 & 0.897 & 0.683 & 0.626 & 0.913 & 0.686 \\
      & VQAv2 & 0.466 & 0.809 & 0.576 & 0.454 & 0.809 & 0.561 \\
      & RealworldQA & 0.538 & 0.810 & 0.664 & 0.517 & 0.817 & 0.632 \\
    \midrule
    \multirow{3}{*}{Gemma3-12B}
      & DocVQA & 0.737 & 0.856 & 0.860 & 0.682 & 0.881 & 0.775 \\
      & VQAv2 & 0.597 & 0.749 & 0.797 & 0.589 & 0.772 & 0.764 \\
      & RealworldQA & 0.639 & 0.712 & 0.898 & 0.592 & 0.689 & 0.860 \\
    \bottomrule
  \end{tabular}
  \caption{\textbf{Frozen and refitted modality weights.} Example-level linear cue-combination weights underlying Tables~\ref{tab:degradation-transfer} and~\ref{tab:degradation-transfer-large}. Frozen weights are fitted on the text noise $\times$ blur boundary and transferred unchanged; refitted weights are estimated on the relevance $\times$ occlusion boundary, using the same example partitions and fitting procedure. Ratios are computed before rounding. Across all $21$ VLM--dataset pairs, the absolute relative ratio change, $|r_{\mathrm{refit}}/r_{\mathrm{frozen}}-1|$ with $r=\beta_T/\beta_V$, has a median of $4.5\%$ and a maximum below $10\%$.}
  \label{tab:transfer-refit-weights}
\end{table*}

\FloatBarrier
\Needspace{.9\textheight}
\subsection{Cue Weights and Mean Evidence}\label{app:cue-components}
Table~\ref{tab:cue-components-ood} reports the individual weights and mean evidence underlying the ratios in Table~\ref{tab:bias-decomposition}, together with the mean and standard deviation of the prior in Eq.~\ref{eq:cue_combination}.
For the cell-level fit, we average the input probes and joint score across training examples separately within each of the $11$ boundary cells, then fit the same nonnegative two-weight model by least squares, giving each cell equal weight.
The evidence and prior summaries use clean inputs ($\sigma=0$, $p=0$) from the same held-out examples as Table~\ref{tab:bias-decomposition}; these examples are not used to estimate either set of weights.
Cell-level fitting describes population mean preferences and does not by itself identify a correction for measurement error in the individual probes.

\begin{table*}[!htbp]
  \centering
  \small
  \setlength{\tabcolsep}{3pt}
  \begin{tabular}{@{}llrrrrrrrr@{}}
    \toprule
    \multirow{2}{*}{Model} & \multirow{2}{*}{Dataset} & \multicolumn{2}{c}{Example-level} & \multicolumn{2}{c}{Cell-level [95\% CI]} & \multicolumn{2}{c}{Mean evidence} & \multicolumn{2}{c}{Prior} \\
    \cmidrule(lr){3-4}\cmidrule(lr){5-6}\cmidrule(lr){7-8}\cmidrule(l){9-10}
     & & $\beta_T$ & $\beta_V$ & $\beta_T$ & $\beta_V$ & $\overline e_T$ & $\overline e_V$ & $\overline\pi_0$ & $\mathrm{SD}(\pi_0)$ \\
    \midrule
    \multirow{3}{*}{Qwen2.5-VL-3B} & DocVQA & $0.831$ & $0.782$ & \shortstack[r]{$0.973$\\{\scriptsize$[0.960, 0.987]$}} & \shortstack[r]{$0.985$\\{\scriptsize$[0.956, 1.018]$}} & $3.739$ & $-2.747$ & $0.052$ & $1.958$ \\[2pt]
     & VQAv2 & $0.855$ & $0.590$ & \shortstack[r]{$1.068$\\{\scriptsize$[1.045, 1.093]$}} & \shortstack[r]{$0.840$\\{\scriptsize$[0.803, 0.878]$}} & $4.780$ & $-3.422$ & $0.041$ & $2.463$ \\[2pt]
     & RealworldQA & $0.841$ & $0.586$ & \shortstack[r]{$0.993$\\{\scriptsize$[0.965, 1.019]$}} & \shortstack[r]{$0.909$\\{\scriptsize$[0.846, 0.985]$}} & $3.205$ & $-1.203$ & $0.044$ & $1.820$ \\[2pt]
    \midrule
    \multirow{3}{*}{Qwen3.5-4B} & DocVQA & $0.542$ & $0.820$ & \shortstack[r]{$0.699$\\{\scriptsize$[0.679, 0.721]$}} & \shortstack[r]{$1.074$\\{\scriptsize$[1.041, 1.109]$}} & $4.357$ & $-3.029$ & $0.168$ & $1.020$ \\[2pt]
     & VQAv2 & $0.426$ & $0.756$ & \shortstack[r]{$0.545$\\{\scriptsize$[0.524, 0.568]$}} & \shortstack[r]{$0.940$\\{\scriptsize$[0.908, 0.978]$}} & $6.130$ & $-3.012$ & $-0.152$ & $1.105$ \\[2pt]
     & RealworldQA & $0.477$ & $0.707$ & \shortstack[r]{$0.576$\\{\scriptsize$[0.554, 0.600]$}} & \shortstack[r]{$0.915$\\{\scriptsize$[0.872, 0.963]$}} & $4.768$ & $-1.653$ & $-0.102$ & $0.739$ \\[2pt]
    \midrule
    \multirow{3}{*}{Gemma3-4B} & DocVQA & $0.724$ & $0.751$ & \shortstack[r]{$0.886$\\{\scriptsize$[0.865, 0.909]$}} & \shortstack[r]{$0.998$\\{\scriptsize$[0.955, 1.044]$}} & $8.212$ & $-5.208$ & $-0.552$ & $5.162$ \\[2pt]
     & VQAv2 & $0.699$ & $0.670$ & \shortstack[r]{$1.061$\\{\scriptsize$[1.017, 1.104]$}} & \shortstack[r]{$1.031$\\{\scriptsize$[0.970, 1.093]$}} & $12.034$ & $-8.531$ & $0.691$ & $6.079$ \\[2pt]
     & RealworldQA & $0.772$ & $0.610$ & \shortstack[r]{$0.947$\\{\scriptsize$[0.902, 0.994]$}} & \shortstack[r]{$1.000$\\{\scriptsize$[0.898, 1.121]$}} & $7.272$ & $-1.930$ & $-0.488$ & $3.669$ \\[2pt]
    \midrule
    \multirow{3}{*}{GLM-4.6V-Flash} & DocVQA & $0.603$ & $0.928$ & \shortstack[r]{$0.741$\\{\scriptsize$[0.720, 0.761]$}} & \shortstack[r]{$1.161$\\{\scriptsize$[1.129, 1.195]$}} & $6.273$ & $-3.374$ & $-0.141$ & $2.541$ \\[2pt]
     & VQAv2 & $0.479$ & $0.773$ & \shortstack[r]{$0.660$\\{\scriptsize$[0.632, 0.687]$}} & \shortstack[r]{$1.084$\\{\scriptsize$[1.027, 1.139]$}} & $8.343$ & $-2.948$ & $-0.239$ & $3.447$ \\[2pt]
     & RealworldQA & $0.455$ & $0.815$ & \shortstack[r]{$0.598$\\{\scriptsize$[0.570, 0.628]$}} & \shortstack[r]{$1.121$\\{\scriptsize$[1.063, 1.186]$}} & $5.808$ & $-1.890$ & $0.076$ & $2.562$ \\[2pt]
    \bottomrule
  \end{tabular}
  \caption{\textbf{Weights, evidence, and prior in linear cue combination.} Example-level weights are the saved OOD fits. Cell-level weights fit the same nonnegative two-weight model to the $11$ training boundary-cell means with equal weight per cell, fixed prior coefficient one, and no intercept. Brackets give pointwise 95\% percentile intervals from $2{,}000$ whole-example refit bootstraps (Appendix~\ref{app:weight-bootstrap}). The evidence means and prior mean and standard deviation are computed on clean held-out test inputs ($\sigma=0$, $p=0$), matching the evidence ratios in Table~\ref{tab:bias-decomposition}. Standard deviations use divisor $N$, the number of held-out examples. Input probes are clipped to $[-30,30]$ before subtracting the prior. Gemma uses its measured knowledge-prefilled prior and answer-prefilled evidence probes.}
  \label{tab:cue-components-ood}
\end{table*}

\FloatBarrier

\clearpage
\section{Supplementary Evaluations}\label{app:supplementary-evaluations}

\FloatBarrier
\subsection{In-Distribution Predictor Comparison}\label{app:id-comparison}

Table~\ref{tab:predictor-comparison-id} repeats Table~\ref{tab:predictor-comparison} with all $36$ degradation cells represented in training, validation, and testing, while keeping the underlying example IDs disjoint. This comparison tests whether relaxing the linear cue-combination constraints or adding a bilinear interaction improves prediction when the full corruption grid is represented during fitting.

\begin{table}[!htbp]
  \centering
  \small
  \setlength{\tabcolsep}{4pt}
  \begin{tabular}{llcc}
    \toprule
    Model & Predictor & KL $\downarrow$ & $R^2$ $\uparrow$ \\
    \midrule
    \multirow{5}{*}{Qwen2.5-VL-3B} & Best Single Probe & 0.242 & 0.494 \\
     & Calibrated Naive Bayes & 0.121 & 0.695 \\
     & Linear Cue Combination & \textbf{0.073} & \textbf{0.827} \\
    \cmidrule(lr){2-4}
     & Bilinear Cue Combination & 0.074 & 0.826 \\
     & Unconstrained Linear & 0.074 & 0.828 \\
    \midrule
    \multirow{5}{*}{Qwen3.5-4B} & Best Single Probe & 0.364 & 0.192 \\
     & Calibrated Naive Bayes & 0.188 & 0.379 \\
     & Linear Cue Combination & \textbf{0.089} & \textbf{0.794} \\
    \cmidrule(lr){2-4}
     & Bilinear Cue Combination & 0.088 & 0.795 \\
     & Unconstrained Linear & 0.080 & 0.809 \\
    \midrule
    \multirow{5}{*}{Gemma3-4B} & Best Single Probe & 0.946 & 0.416 \\
     & Calibrated Naive Bayes & 0.612 & 0.501 \\
     & Linear Cue Combination & \textbf{0.364} & \textbf{0.672} \\
    \cmidrule(lr){2-4}
     & Bilinear Cue Combination & 0.365 & 0.670 \\
     & Unconstrained Linear & 0.373 & 0.674 \\
    \midrule
    \multirow{5}{*}{GLM-4.6V-Flash} & Best Single Probe & 0.551 & 0.467 \\
     & Calibrated Naive Bayes & 0.309 & 0.571 \\
     & Linear Cue Combination & \textbf{0.113} & \textbf{0.845} \\
    \cmidrule(lr){2-4}
     & Bilinear Cue Combination & 0.113 & 0.846 \\
     & Unconstrained Linear & 0.115 & 0.850 \\
    \bottomrule
  \end{tabular}
  \caption{ID replication of Table~\ref{tab:predictor-comparison}, averaging the three dataset-level metrics equally. KL is the mean divergence from the observed to the predicted two-answer distribution; $R^2$ compares predicted and observed joint log-odds. Bold marks the best of the three predictors above the inner rule.}
  \label{tab:predictor-comparison-id}
\end{table}

\FloatBarrier
\Needspace{.55\textheight}
\subsection{Random Forest Results}\label{app:random-forest}

We evaluate a random forest as a flexible nonlinear predictor of the joint log-odds from the same unimodal probes and prior.
Fitting and hyperparameter selection follow Appendix~\ref{app:predictor-fitting}, with the same example-level splits as the parametric predictors.
Table~\ref{tab:random-forest-comparison} compares it with linear cue combination under the ID protocol for the four main-paper VLMs, with all $36$ degradation cells represented during fitting.
The random forest improves mean $R^2$ over linear cue combination by at most $0.015$.

\begin{table}[!htbp]
  \centering
  \small
  \setlength{\tabcolsep}{4pt}
  \begin{tabular}{llcc}
    \toprule
    Model & Predictor & KL $\downarrow$ & $R^2$ $\uparrow$ \\
    \midrule
    \multirow{2}{*}{Qwen2.5-VL-3B} & Linear Cue Combination & \textbf{0.073} & \textbf{0.827} \\
     & Random Forest & 0.077 & 0.820 \\
    \midrule
    \multirow{2}{*}{Qwen3.5-4B} & Linear Cue Combination & 0.089 & 0.794 \\
     & Random Forest & \textbf{0.082} & \textbf{0.809} \\
    \midrule
    \multirow{2}{*}{Gemma3-4B} & Linear Cue Combination & \textbf{0.364} & 0.672 \\
     & Random Forest & 0.421 & \textbf{0.674} \\
    \midrule
    \multirow{2}{*}{GLM-4.6V-Flash} & Linear Cue Combination & \textbf{0.113} & \textbf{0.845} \\
     & Random Forest & 0.134 & 0.830 \\
    \bottomrule
  \end{tabular}
  \caption{ID comparison of random forest and linear cue combination, using the same protocol as Table~\ref{tab:predictor-comparison-id} and averaging the three dataset-level metrics equally. KL is the mean divergence from the observed to the predicted two-answer distribution; $R^2$ compares predicted and observed joint log-odds. Bold marks the better predictor within each VLM and metric.}
  \label{tab:random-forest-comparison}
\end{table}

\FloatBarrier
\Needspace{.72\textheight}
\subsection{Larger-Model Replications}\label{app:larger-models}

We repeat the predictor comparison and weight analysis for Qwen2.5-VL-7B, Qwen3.5-9B, and Gemma3-12B. 
We do not evaluate the larger GLM-4.6V because its $108$B-parameter checkpoint is substantially larger than the other models considered here.
Table~\ref{tab:predictor-comparison-large} replicates Table~\ref{tab:predictor-comparison} under the same OOD protocol; Table~\ref{tab:fitted-params-large-ood} extends the example-level weights in Table~\ref{tab:cue-components-ood} to the larger models.

\begin{table}[!htbp]
  \centering
  \small
  \setlength{\tabcolsep}{4pt}
  \begin{tabular}{llcc}
    \toprule
    Model & Predictor & KL $\downarrow$ & $R^2$ $\uparrow$ \\
    \midrule
    \multirow{5}{*}{Qwen2.5-VL-7B} & Best Single Probe & 0.231 & 0.591 \\
     & Calibrated Naive Bayes & 0.116 & 0.732 \\
     & Linear Cue Combination & \textbf{0.064} & \textbf{0.865} \\
    \cmidrule(lr){2-4}
     & Bilinear Cue Combination & 0.064 & 0.866 \\
     & Unconstrained Linear & 0.065 & 0.865 \\
    \midrule
    \multirow{5}{*}{Qwen3.5-9B} & Best Single Probe & 0.094 & 0.272 \\
     & Calibrated Naive Bayes & 0.056 & 0.506 \\
     & Linear Cue Combination & \textbf{0.022} & \textbf{0.820} \\
    \cmidrule(lr){2-4}
     & Bilinear Cue Combination & 0.022 & 0.819 \\
     & Unconstrained Linear & 0.021 & 0.821 \\
    \midrule
    \multirow{5}{*}{Gemma3-12B} & Best Single Probe & 0.717 & 0.506 \\
     & Calibrated Naive Bayes & 0.499 & 0.568 \\
     & Linear Cue Combination & \textbf{0.263} & \textbf{0.763} \\
    \cmidrule(lr){2-4}
     & Bilinear Cue Combination & 0.268 & 0.760 \\
     & Unconstrained Linear & 0.278 & 0.761 \\
    \bottomrule
  \end{tabular}
  \caption{OOD predictor comparison for the three larger VLMs, replicating Table~\ref{tab:predictor-comparison}. Metrics average the three datasets equally. KL is the mean divergence from the observed to the predicted two-answer distribution; $R^2$ compares predicted and observed joint log-odds. Bold marks the best of the three predictors above the inner rule.}
  \label{tab:predictor-comparison-large}
\end{table}

\begin{table*}[!htbp]
  \centering
  \small
  \setlength{\tabcolsep}{3pt}
  \begin{tabular}{@{}llrrrr@{}}
    \toprule
    \multirow{2}{*}{Model} & \multirow{2}{*}{Dataset} & \multicolumn{2}{c}{Example-level} & \multicolumn{2}{c}{Cell-level [95\% CI]} \\
    \cmidrule(lr){3-4}\cmidrule(lr){5-6}
     & & $\beta_T$ & $\beta_V$ & $\beta_T$ & $\beta_V$ \\
    \midrule
    \multirow{3}{*}{Qwen2.5-VL-7B} & DocVQA & $0.801$ & $0.824$ & \shortstack[r]{$0.929$\\{\scriptsize$[0.914, 0.944]$}} & \shortstack[r]{$0.996$\\{\scriptsize$[0.972, 1.024]$}} \\[2pt]
     & VQAv2 & $0.695$ & $0.698$ & \shortstack[r]{$0.846$\\{\scriptsize$[0.820, 0.873]$}} & \shortstack[r]{$0.854$\\{\scriptsize$[0.823, 0.887]$}} \\[2pt]
     & RealworldQA & $0.732$ & $0.714$ & \shortstack[r]{$0.871$\\{\scriptsize$[0.837, 0.901]$}} & \shortstack[r]{$0.995$\\{\scriptsize$[0.945, 1.055]$}} \\[2pt]
    \midrule
    \multirow{3}{*}{Qwen3.5-9B} & DocVQA & $0.613$ & $0.897$ & \shortstack[r]{$0.725$\\{\scriptsize$[0.711, 0.740]$}} & \shortstack[r]{$1.085$\\{\scriptsize$[1.060, 1.113]$}} \\[2pt]
     & VQAv2 & $0.466$ & $0.809$ & \shortstack[r]{$0.572$\\{\scriptsize$[0.548, 0.595]$}} & \shortstack[r]{$1.014$\\{\scriptsize$[0.977, 1.054]$}} \\[2pt]
     & RealworldQA & $0.538$ & $0.810$ & \shortstack[r]{$0.615$\\{\scriptsize$[0.593, 0.638]$}} & \shortstack[r]{$1.084$\\{\scriptsize$[1.035, 1.135]$}} \\[2pt]
    \midrule
    \multirow{3}{*}{Gemma3-12B} & DocVQA & $0.737$ & $0.856$ & \shortstack[r]{$0.920$\\{\scriptsize$[0.897, 0.943]$}} & \shortstack[r]{$1.113$\\{\scriptsize$[1.068, 1.160]$}} \\[2pt]
     & VQAv2 & $0.597$ & $0.749$ & \shortstack[r]{$0.856$\\{\scriptsize$[0.826, 0.887]$}} & \shortstack[r]{$1.139$\\{\scriptsize$[1.078, 1.205]$}} \\[2pt]
     & RealworldQA & $0.639$ & $0.712$ & \shortstack[r]{$0.803$\\{\scriptsize$[0.772, 0.834]$}} & \shortstack[r]{$1.181$\\{\scriptsize$[1.087, 1.281]$}} \\[2pt]
    \bottomrule
  \end{tabular}
  \caption{\textbf{Fitted cue weights for larger VLMs.} Example-level weights are the saved OOD fits. Cell-level weights fit the same nonnegative two-weight model to the $11$ training boundary-cell means with equal weight per cell, fixed prior coefficient one, and no intercept. Brackets give pointwise 95\% percentile intervals from $2{,}000$ whole-example refit bootstraps (Appendix~\ref{app:weight-bootstrap}). Gemma uses its measured knowledge-prefilled prior and answer-prefilled evidence probes.}
  \label{tab:fitted-params-large-ood}
\end{table*}

\FloatBarrier
\paragraph{Integration bias and evidence ratio.}
Table~\ref{tab:bias-decomposition-large} repeats Table~\ref{tab:bias-decomposition} for the three larger models, comparing example-level and cell-level weight ratios alongside evidence ratios on clean inputs from the same held-out test examples.

\begin{table*}[!htbp]
  \centering
  \small
  \setlength{\tabcolsep}{4pt}
  \begin{tabular}{@{}llccc@{}}
    \toprule
    & & \multicolumn{2}{c}{$\beta_T/\beta_V$ [95\% CI]} & $|\bar e_T|/|\bar e_V|$ \\
    \cmidrule(lr){3-4}\cmidrule(lr){5-5}
    Model & Dataset & Example & Cell & Clean \\
    \midrule
    \multirow{3}{*}{Qwen2.5-VL-7B} & DocVQA & $0.972\;[0.939, 1.003]$ & $0.933\;[0.901, 0.964]$ & 1.20 \\
     & VQAv2 & $0.995\;[0.952, 1.040]$ & $0.991\;[0.950, 1.030]$ & 1.31 \\
     & RealworldQA & $1.025\;[0.965, 1.087]$ & $0.876\;[0.818, 0.935]$ & 2.42 \\
    \midrule
    \multirow{3}{*}{Qwen3.5-9B} & DocVQA & $0.683\;[0.667, 0.709]$ & $0.668\;[0.646, 0.690]$ & 1.46 \\
     & VQAv2 & $0.576\;[0.552, 0.630]$ & $0.564\;[0.538, 0.591]$ & 2.01 \\
     & RealworldQA & $0.664\;[0.613, 0.687]$ & $0.568\;[0.534, 0.604]$ & 3.39 \\
    \midrule
    \multirow{3}{*}{Gemma3-12B} & DocVQA & $0.860\;[0.820, 0.912]$ & $0.826\;[0.786, 0.869]$ & 1.29 \\
     & VQAv2 & $0.797\;[0.750, 0.828]$ & $0.752\;[0.710, 0.795]$ & 1.85 \\
     & RealworldQA & $0.898\;[0.834, 0.968]$ & $0.680\;[0.620, 0.743]$ & 3.03 \\
    \bottomrule
  \end{tabular}
  \caption{\textbf{Integration bias and evidence ratio in larger VLMs.} Ratios above one indicate greater text weighting or stronger text evidence. Brackets give pointwise 95\% percentile intervals from $2{,}000$ whole-example refit bootstraps (Appendix~\ref{app:weight-bootstrap}). Example fits repeat validation checkpoint selection; cell fits recompute the $11$ training boundary-cell means in each resample. Evidence ratios are the absolute values of signed mean evidence on clean held-out inputs ($\sigma=0$, $p=0$), with absolute values taken after averaging.}
  \label{tab:bias-decomposition-large}
\end{table*}

\FloatBarrier
\paragraph{Transfer across reliability manipulations.}

Table~\ref{tab:degradation-transfer-large} repeats the weight-transfer comparison for these larger models. We compare weights refitted on relevance $\times$ occlusion against the frozen text noise $\times$ blur weights in Table~\ref{tab:fitted-params-large-ood}, using identical target-grid test examples and the OOD protocol in both cases.

\begin{table}[!htbp]
  \centering
  \small
  \setlength{\tabcolsep}{4pt}
  \begin{tabular}{lrr}
    \toprule
    Model & Refit $R^2$ & Transferred $R^2$ \\
    \midrule
    Qwen2.5-VL-7B & 0.822 & 0.830 \\
    Qwen3.5-9B & 0.759 & 0.757 \\
    Gemma3-12B & 0.688 & 0.689 \\
    \bottomrule
  \end{tabular}
  \caption{Weight transfer for the larger VLMs, replicating Table~\ref{tab:degradation-transfer}. Refit estimates weights on the boundary of relevance $\times$ pixel occlusion; Transferred freezes weights estimated on the boundary of text noise $\times$ blur. Both are evaluated on the same held-out interior examples of the new grid. Values average the three dataset-level $R^2$ scores equally.}
  \label{tab:degradation-transfer-large}
\end{table}

\FloatBarrier

\clearpage
\section{Resources and Disclosures}\label{app:resources}

\FloatBarrier
\subsection{Computational Resources}\label{app:compute}
All experiments were run on a single node with 8 NVIDIA L40S GPUs.
Our experiments are inference-only, we do not train or fine-tune any VLM.
Predictor fitting runs on CPU: the linear, bilinear, and unconstrained models use full-batch L-BFGS, while the random forest uses decision-tree ensembles (Appendix~\ref{app:predictor-fitting}).

\FloatBarrier
\subsection{Models and Checkpoints}\label{app:models}
Table~\ref{tab:model-checkpoints} lists the checkpoint identifiers used for the four main VLMs and the larger-model replications in Appendix~\ref{app:larger-models}.
Each VLM uses its native processor and chat template; scoring conventions, including Gemma3's assistant prefixes, are given in Appendix~\ref{app:behavioral-measurements}.
\begin{table}[!htbp]
  \centering
  \small
  \begin{tabular}{lll}
    \toprule
    Evaluation & VLM & Checkpoint identifier \\
    \midrule
    Main & Qwen2.5-VL-3B & \texttt{Qwen/Qwen2.5-VL-3B-Instruct} \\
    Main & Qwen3.5-4B & \texttt{Qwen/Qwen3.5-4B} \\
    Main & Gemma3-4B & \texttt{google/gemma-3-4b-it} \\
    Main & GLM-4.6V-Flash & \texttt{zai-org/GLM-4.6V-Flash} \\
    \midrule
    Larger & Qwen2.5-VL-7B & \texttt{Qwen/Qwen2.5-VL-7B-Instruct} \\
    Larger & Qwen3.5-9B & \texttt{Qwen/Qwen3.5-9B} \\
    Larger & Gemma3-12B & \texttt{google/gemma-3-12b-it} \\
    \bottomrule
  \end{tabular}
  \caption{Checkpoint identifiers for the VLMs evaluated in this paper. Gemma3 checkpoints are loaded from local copies of the listed repositories.}
  \label{tab:model-checkpoints}
\end{table}

\FloatBarrier
\subsection{Artifact Licensing and Intended Use}\label{app:licensing}
The source benchmarks we build on are publicly released for research use: DocVQA \citep{mathew2021doc}, VQAv2 \citep{antol2015vqa}, and RealworldQA \citep{xai2024grok}, together with the conflicting-text source of \citet{deng2025blind}.
The four models we study, Qwen2.5-VL \citep{qwen2.5-VL}, Qwen3.5 \citep{qwen3.5}, Gemma3 \citep{Kamath2025Gemma3}, and GLM-4.6V \citep{glm45v}, are distributed as open-weight models under their respective licenses.
We use all of these artifacts solely for non-commercial research, consistent with their intended use as research benchmarks and openly released models.

\FloatBarrier
\subsection{Release Information}\label{app:release}
Code and data will be released upon publication.

\FloatBarrier
\subsection{Usage of AI Tools}\label{app:ai-tools}
AI tools were used to help write and debug code for the experiments in this project and to polish the prose of the paper draft.
All experimental design, analysis, and claims are the authors' own, and the authors take full responsibility for the content of the paper.

\FloatBarrier

\end{document}